\documentclass{article}
\usepackage{arxiv}
\usepackage[utf8]{inputenc}
\usepackage[T1]{fontenc}
\usepackage{amsmath}
\usepackage{amssymb}
\usepackage{amsthm, amsfonts,bm}
\usepackage{mathtools}
\usepackage{supertabular}
\usepackage{graphicx}
\usepackage{caption}
\usepackage{subcaption}
\usepackage{comment}
\usepackage{listings}
\usepackage{xcolor}
\usepackage{float}
\usepackage{cases}
\usepackage{bigints}
\usepackage{enumerate}
\usepackage{algorithm2e}
\usepackage{algpseudocode}
\usepackage{hyperref}
\usepackage{mathtools}
\usepackage{bbm}
\usepackage{amsmath}
\usepackage[capitalise]{cleveref}
\usepackage{array}
\usepackage{url}
\usepackage{makecell}

\title{Mixture-of-Experts Graph Transformers for Interpretable Particle Collision Detection}

\author{
 Donatella Genovese \\
  Sapienza University of Rome\\
  \texttt{donatella.genovese@uniroma1.it} \\
   \And
 Alessandro Sgroi \\
  Sapienza University of Rome\\
  \texttt{sgroi.1658835@studenti.uniroma1.it } \\
  \And
 Alessio Devoto \\
  Sapienza University of Rome\\
  \texttt{alessio.devoto@uniroma1.it} \\
  \And
   Samuel Valentine \\
  University of Liverpool\\
  \texttt{S.Valentine@liverpool.ac.uk} \\
  \And
     Lennox Wood \\
  University of Liverpool\\
  \texttt{sglwoo10@liverpool.ac.uk} \\
  \And 
     Cristiano Sebastiani \\
  Experimental Physics Department, CERN\\
  \texttt{cristiano.sebastiani@cern.ch} \\
  \And 
     Stefano Giagu \\
  Sapienza University of Rome\\
  \texttt{stefano.giagu@uniroma1.it} \\
  \And
     Monica D'Onofrio \\
  University of Liverpool\\
  \texttt{onofrio@liverpool.ac.uk} \\
  \And
       Simone Scardapane \\
  Sapienza University of Rome\\
  \texttt{simone.scardapane@uniroma1.it} 
}
\newcommand{\particle}[1]{\textit{#1}}

\begin{document}

\maketitle

\begin{abstract}
The Large Hadron Collider at CERN produces immense volumes of complex data from high-energy particle collisions, demanding sophisticated analytical techniques for effective interpretation. Neural Networks, including Graph Neural Networks, have shown promise in tasks such as event classification and object identification by representing collisions as graphs. However, while Graph Neural Networks excel in predictive accuracy, their "black box" nature often limits their explainability, making it difficult to trust their decision-making processes.
In this paper, we propose a novel approach that combines a Graph Transformer model with Mixture-of-Expert layers to achieve high predictive performance while embedding explainability into the architecture. By leveraging attention maps and expert specialization, the model offers insights into its internal decision-making, linking predictions to physics-informed features. We evaluate the model on simulated events from the ATLAS experiment, focusing on distinguishing rare Supersymmetric signal events from Standard Model background.
Our results highlight that the model achieves competitive classification accuracy while providing interpretable outputs that align with known physics, demonstrating its potential as a robust and transparent tool for high-energy physics data analysis. This approach underscores the importance of explainability in machine learning methods applied to high energy physics, offering a path toward greater trust in AI-driven discoveries.


\end{abstract}

\section*{Introduction}

The Large Hadron Collider (LHC) at CERN offers an exceptional setting for investigating the fundamental constituents of matter. Experiments like ATLAS \cite{ATLAS}, generate extensive data streams from high-energy particle collisions, enabling the study of particle interactions across a wide energy spectrum. While this data is invaluable for advancing particle physics, the sheer scale and complexity of the information present significant analytical challenges \cite{LHC}. For instance, ATLAS event reconstruction processes data at rates exceeding 3.5 terabytes per second \cite{seconds}, requiring advanced computational techniques to extract meaningful insights.

In this work, we leverage simulated LHC data as a controlled benchmark to address a key challenge in modern high-energy physics (HEP): the development of interpretable machine-learning models. Simulated data, where the underlying physics is fully known, provides an ideal environment to evaluate the trustworthiness of machine-learning methods by verifying whether their outputs align with established physical principles. Machine learning (ML) and deep learning techniques have demonstrated strong potential in tackling HEP challenges \cite{Challenges}, including particle identification \cite{particle_identification}, event reconstruction \cite{event_reconstraction}, and background subtraction \cite{background}. Nevertheless, these methods often operate as "black boxes", making it difficult to interpret their decision-making processes. This lack of transparency is a critical limitation, particularly in fields like HEP, where explainability is essential for establishing confidence in novel discoveries. 

From a performance point of view, Graph Neural Networks (GNNs) have emerged as promising tools for analyzing complex datasets like those produced by the LHC. By representing particles and their interactions as nodes and edges in a graph \cite{GNNPhys}, GNNs can model intricate relationships that other methods might overlook.  While GNNs have demonstrated state-of-the-art performance in classification tasks, their value extends beyond predictive accuracy. A major focus of this work is on developing GNN-based models that inherently provide insights into their internal decision-making processes, ensuring that their predictions align with physical intuition. 
The terms \emph{explainability} and \emph{interpretability} are often used interchangeably in the literature, although some works attempt to distinguish between them. However, there is currently no universally accepted consensus within the Explainable Artificial Intelligence (XAI) community on how to clearly differentiate these concepts \cite{EX1} \cite{EX2} \cite{EX3} \cite{EX4}. Given this ongoing debate and lack of standardization, in this work we adopt the term explainability to refer broadly to the overall process of making AI systems more understandable to humans.

Despite advances in these areas, GNNs often remain "black box" models. To address this issue, various post-hoc explainability methods have been introduced for graph neural networks \cite{XAIGRAPH}, such as PGExplainer\cite{PGEx} and GNNExplainer \cite{GNNEx}, alongside data attribution techniques \cite{GNNDataAttr}. These approaches aim to provide insight into how GNNs make their predictions by identifying important input features or graph substructures. Nonetheless, while these methods offer some understanding, they still fall short of providing a comprehensive understanding of the model's internal workings \cite{FragInt,SanityCheck}. 

Our work addresses these limitations by proposing a novel model that integrates explainability directly into its architecture. Specifically, we build upon the Graph Transformer (GT) model \cite{GraphTransformer}. This architecture offers the possibility of visualizing attention maps, enabling us to illustrate how the model identifies relationships within graph structures.
By examining these attention maps, we can verify whether the model focuses on graph regions consistent with known physics. To further enhance explainability, we replace the standard feed-forward layer of the Transformer with a Mixture of Experts (MoE) layer \cite{MoE}. This design enables the model to distribute its outputs across specialized "experts," each targeting distinct data subsets, thereby providing a clearer understanding of how individual components contribute to the final decision.
This enhanced transparency allows us to trace the model’s reasoning pathways and verify their alignment with physical principles.
We test our approach on Monte Carlo-simulated collision events, targeting hypothetical particles predicted by Supersymmetry (SUSY). These datasets include both rare SUSY signal events and dominant Standard Model (SM) backgrounds, following established ATLAS methodologies \cite{Susy}. The classification task, distinguishing rare signal from high-rate background, mirrors the challenges typical in HEP. Our focus is not solely on achieving high performance but on validating the explainability of the model. By leveraging this controlled dataset, we demonstrate how our method identifies critical features and regions of the graph that align with the expected physics, fostering trust in the model’s predictions.
In this work, we emphasize that the explainability of machine-learning models is as crucial as their accuracy, particularly when exploring the potential for new physics. 
Our approach illustrates how explainable GNN-based methods can provide high predictive accuracy, reinforcing the connection between computational predictions and physical intuition. This focus on explainability ensures that our methodology is not only a powerful analytical tool but also a reliable framework for future discoveries in particle physics.


\section{Related Works} 

\subsection{Graph Neural Networks and Graph Transformers}
GNNs are a specialized class of neural networks designed to process and analyze graph-structured data, with their key strength lying in their ability to learn representations that capture complex relationships among nodes in a graph. GNNs are extremely versatile and can be adapted to various graph-related tasks, including node and graph classification, regression, edge prediction, graph generation, and node clustering. They have found wide-ranging applications in particle physics \cite{GNNPhys}, \cite{GNNPhys2} and HEP, especially in particle tracking and reconstruction \cite{GNNPartRec,GNNPartRec2}. The physics tasks at the LHC experiments have provided many potential applications where graph neural networks have been successfully applied \cite{GNNLHC}.
Building upon the foundation of GNNs, GTs have recently emerged as a powerful graph learning method. GTs draw inspiration from the success of Transformers in natural language processing (NLP) \cite{AttentionIsAll} and computer vision (CV) \cite{ViT}. GTs excel at handling dynamic and heterogeneous graphs, utilizing both node and edge features \cite{GraphTransformer}. Various adaptations and expansions of GTs have demonstrated their superiority in addressing multiple challenges in graph learning, such as processing large-scale graphs \cite{GraphLearning}. In addition to these advancements, explainability methods for attention mechanisms have provided valuable insights into the inner workings of self-attention models, further enhancing their explainability \cite{AttVis}.

\subsection{Mixture of Experts}

The Mixture of Experts (MoE, \cite{MoE}) architecture offers a modular approach to scaling neural networks size. 
As models increase in size and complexity, they require substantial computational resources. MoE addresses this challenge by introducing expert specialization, where multiple subnetworks, or "experts", are trained to handle different regions of the input space. Instead of using a single, large model for all inputs, MoE incorporates a gating mechanism that dynamically assigns each input to a subset of experts, selecting the most relevant ones based on learned patterns \cite{MoE}.
This mechanism allows for task-specific specialization, allowing each expert to focus on different aspects of input data \cite{MoEfact}. As a result, the architecture can scale more efficiently, allowing for higher model capacity without the need for proportionally increased computation for every input. By activating only a small number of experts at a time, MoE provides a computationally efficient solution for handling large and complex models \cite{Gshard}. Moreover, recent advancements in explainability for MoE architectures, such as analyzing expert utilization and routing behavior, have deepened our understanding of their decision-making processes and modality-specific specializations\cite{MoEXAI}.

\subsection{Explainability for High Energy Physics}

Several explainability methods have been applied to HEP tasks, illustrating their utility across different domains. Models like XGBoost \cite{XGBOOST} and Deep Neural Networks (DNNs) have been interpreted using SHapley Additive exPlanations (SHAP) \cite{SHAP}. SHAP studies have proven effective to understand how signal events, characterised by the presence of a Higgs boson candidate, are distinguished from background noise through physically relevant kinematic variables \cite{SHAPHEP}. Layerwise Relevance Propagation (LRP)\cite{LRP} has been used to enhance the explainability of DNNs for classifying boosted Z boson jets \cite{LRPJETTAG}. Through LRP-based heatmaps, the method highlighted relevant detector regions. The application of LRP has also been extended to GNNs in the particle-flow reconstruction task \cite{LRPGNNPF}. By distributing relevance scores, the study quantified the importance of input features such as charge and energy, as well as the connections within graph structures. Additionally, data attribution techniques like TracIn\cite{TRACIN} have been employed to identify impactful training data in the ATLAS datasets considered in our paper \cite{Susy} 
This approach improved both model efficiency and explainability by refining training datasets and revealing influential patterns \cite{GNNDataAttr}.
However, these methods often exhibit brittleness and lack reliability in challenging real-world scenarios. For instance, studies have shown that many post hoc explainability techniques fail fundamental sanity checks, raising concerns about their validity and robustness \cite{SanityCheck}. Other research has highlighted the fragility of interpretations, particularly when faced with slight changes in input or model parameters \cite{FragInt}. Attribution-based methods, in particular, have been criticized for being overly sensitive to model biases, making their insights less reliable in practical applications \cite{AttXAI}.
To address these challenges, we propose intrinsic techniques that integrate explainability directly into the model architecture and learning process, enabling the analysis of the model’s internal decision-making mechanisms for complex HEP tasks without relying on post-hoc explanations.

\section{Experimental setup}

\subsection{Dataset}
\label{sec:dataset}
\begin{figure}[!b]
    \centering
    \begin{subfigure}[b]{0.3\textwidth}
         \centering
         \includegraphics[width=\textwidth]{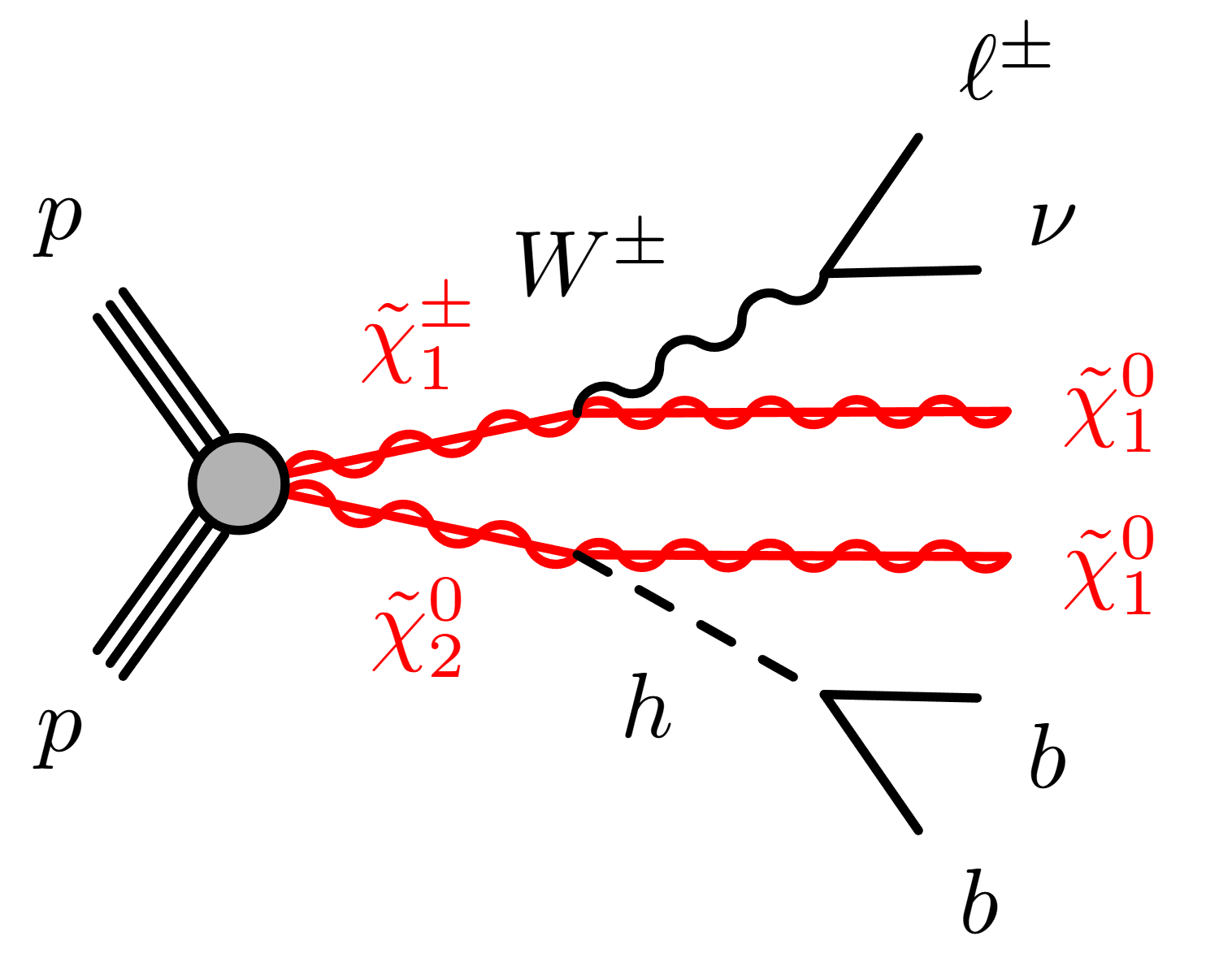}
         \caption{} \label{fig:diagram} 
         \end{subfigure}
         \hspace{3.0cm}
    \begin{subfigure}[b]{0.22\textwidth}
         \centering
        \includegraphics[width=\textwidth]{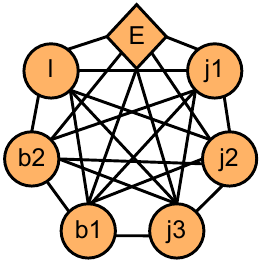}
         \caption{}
         \label{fig:graph}
     \end{subfigure}
     \caption{(a) Diagram of the SUSY signal process, showing chargino and neutralino decaying into W and Higgs bosons, with leptons, neutrinos, and b-quarks in the final state. (b) Schematic representation of a particle collision event modeled as a fully connected graph. Nodes represent reconstructed particles and objects: jets (j1, j2, j3), b-tagged jets (b1, b2), lepton (l), and missing transverse energy (E). The diamond shape for E indicates that it is not a directly reconstructed particle but an inferred quantity.}
\end{figure}

The SUSY dataset, derived from the ATLAS publication \cite{Susy} {and available in the CERN open data repository} (https://opendata.cern.ch/record/28100), contains Monte Carlo simulated collision events for a set of signal models (Signal) and {SM backgrounds -- where the two major sources are hereby considered (Background)}. Details on the two types of events are given below: 
\begin{itemize}
    \item \textbf{Signal}: Collision signal events are depicted in \cref{fig:diagram}. Two parent particles (named chargino, $\tilde{\chi}^{\pm}_{1}$, and next-to-lightest neutralino, $\tilde{\chi}^{0}_{2}$) are produced at the collision point and promptly decay into a W and Higgs boson, respectively, and the dark matter candidate (the lightest-neutralino, $\tilde{\chi}^{0}_{1}$). The W boson decay immediately in a lepton ($l$) and a neutrino ($\nu$), and the Higgs boson in a pair of b-quarks. 
    \item \textbf{Background}: SM background events represent the processes of pair production of two top quarks ($t\bar{t}$) and the production of a top quark and a W boson ($Wt$ or single top). Other sources of SM backgrounds considered in the ATLAS paper are not considered here as potentially more distinguishable from signal and anyway contributing for less than 20\% of the total background. Both top-quark processes mimic the signal as they result in similar final particles. 
\end{itemize}

\begin{figure}[!b]
    \centering
    \begin{subfigure}[b]{0.45\textwidth}
         \centering
        \includegraphics[width=\textwidth]{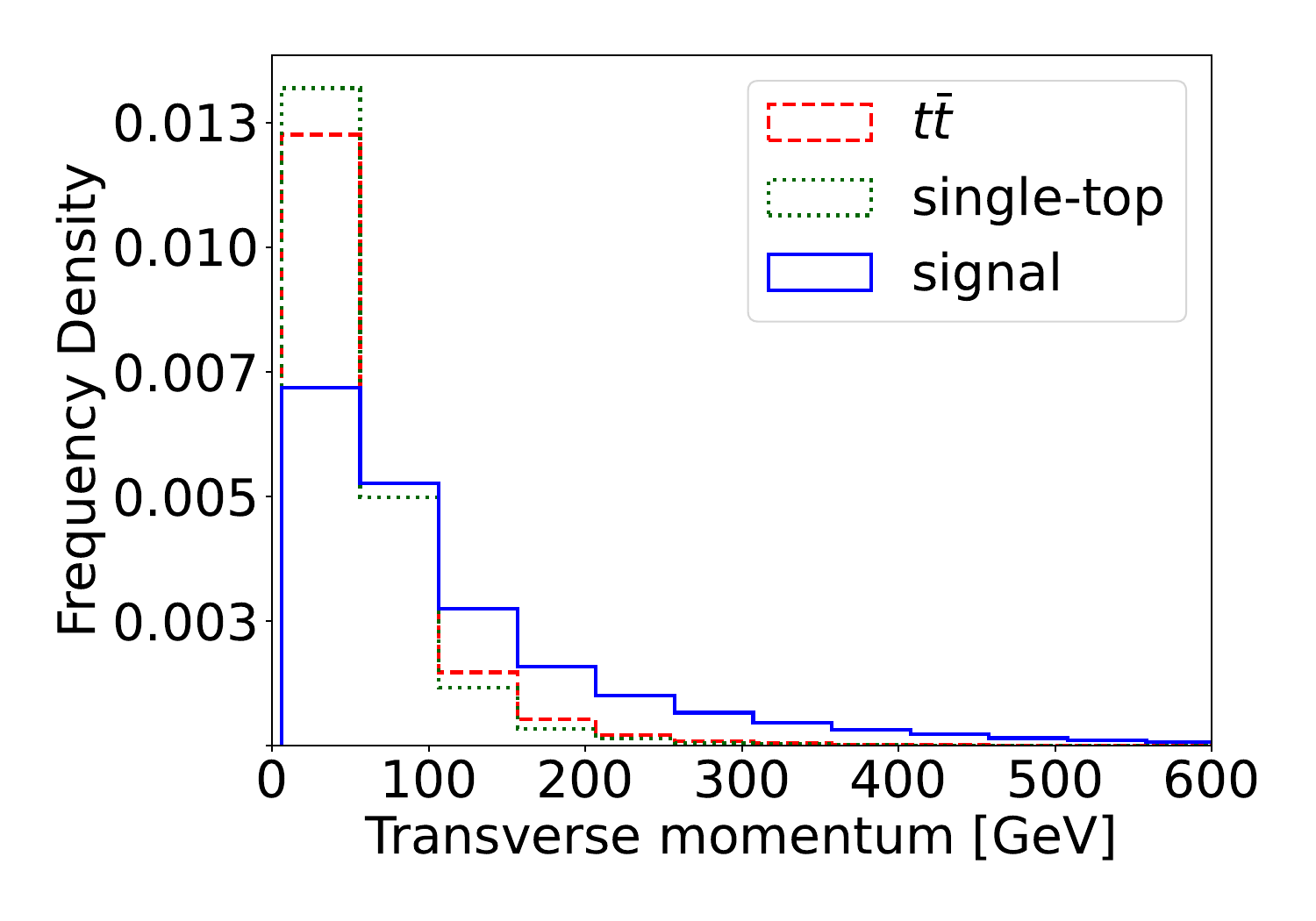}
         \caption{}
         \label{fig:leppt}
     \end{subfigure}
     \hspace{0.5cm}
     \begin{subfigure}[b]{0.45\textwidth}
       \centering
        \includegraphics[width=\textwidth]{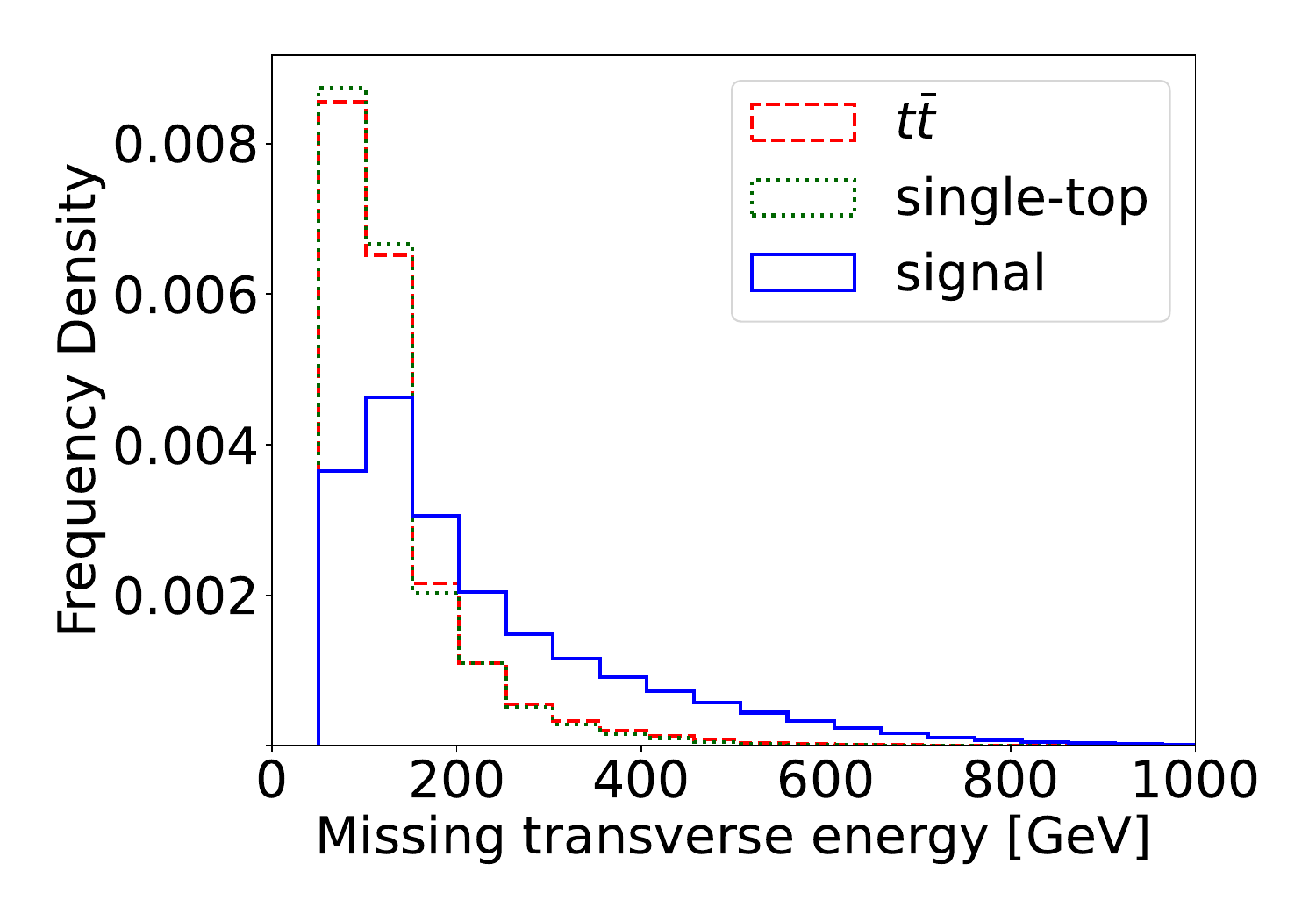}
         \caption{}
         \label{fig:energypt}
     \end{subfigure}
     \caption{Distributions of (a) $p_T$ of lepton and (b) for the $E^{Miss}_{T}$, comparing signal and the two main background processes.}
\label{fig:graphsusy}
\end{figure}

Signal models are parametrised as a function of two variables, the mass of the SUSY parent particles (equal, according to the model) and the mass of the dark matter particle. The mass value ranges between $125$ GeV (the value of the Higgs mass) and 1000 GeV for the former, and 0 and $400$ GeV for the latter, although the highest statistics models are in the region where the mass splitting between parent and dark matter particles is close to the Higgs mass, as directly targeted by the ATLAS paper. The kinematic of the signal events depends on the choices of masses, and each choice represents a different model. If SUSY were realised in nature, only one choice of mass parameters would be the correct one. Since this is unknown a priori, all models are mixed together with equivalent probability.

The main task involves detecting the rare signals amidst a large background of SM processes. In real data, signal events are expected to be well below 0.1\% of the total background, even after dedicated selections are imposed. To distinguish these events, we rely on kinematic features that offer significant discriminative power.  
We represent each collision event as a fully connected graph $G$ with 6-7 nodes ($N$), where each node represents a particle and contains up to 6 kinematic features. Mathematically, each event is encoded as a feature matrix $X \in \mathbb{R}^{N \times 6}$, where rows correspond to particles and columns to their features.
The scheme for particle collisions and their corresponding features are depicted in the graph in \cref{fig:graph} and, in detail, in \cref{tab:kinematic_properties}. With reference to the Table, each row represents a distinct particle that was reconstructed in the detector. The first rows correspond to the jets (\particle{j1, j2, j3}), where jets are produced by the hadronisation process of quarks. The \particle{b1} and \particle{b2} represent the jets originated from the b-quarks present in signal and background events, and are a subset of the generic jets. 
For events with 6 nodes, \particle{j1} and \particle{j2} are equivalent to \particle{b1} and \particle{b2}. Events with 7 nodes are those with an extra jet, not depicted in \cref{fig:diagram}. This categorization is made as the presence of a third jet is expected to offer additional discriminative power.   
The \particle{lepton} row, $l$,  refers to either a muon or an electron, depending on the particle identified. Finally, the \particle{energy} row ($E$) represents the so-called "missing" energy, used as a proxy for particles invisible to the detector such as neutrinos and dark matter candidates. The dataset columns describe various physical properties of the particles. Feature 1, $p_T$, represents the transverse momentum magnitude for particles, while for the \particle{energy} row, it is the $E^{\text{Miss}}_{T}$ quantity (called missing transverse energy). Features 2 and 3 are the pseudorapidity ($\eta$, related to the angle of a particle relative to the beam axis $\theta$ in collision events) and the azimuthal angle ($\phi$) of jet and lepton objects, respectively. Only the azimuthal angle of $E^{\text{Miss}}_{T}$ can be defined due to the unknown longitudinal momentum in head-on collisions at the LHC. Feature 4 is referred to as "quantile" and quantifies the likelihood for a jet to be identified as originating from a b-quark or not according to the ATLAS $b$-tagging identification procedure. Feature 5, $\textit{b1m}$ and $\textit{b2m}$, denote the mass of the b-jets calculated from jet component four-vectors. The final column is $\sigma^{}_{ETMiss}$, which is implemented as a feature of the missing transverse energy quantity and  directly related to its resolution \cite{Susy}. In \cref{fig:graphsusy} (a) and (b), key features are shown for signal and the two kind of background events considered here, normalised to the unit area. Top-pair and single-top backgrounds are added together as kinematically similar. All signal models are mixed together, with models defined by (low) high mass particles contributing mostly to the bulk (tails) of the distributions.

\begin{table*}[!ht]
\centering
\caption{Kinematic properties of particles.}
\label{tab:kinematic_properties}
\begin{tabular}{lcccccc} 
\hline
Particle & F1             & F2           & F3           & F4        & F5         & F6       \\
\hline
\particle{j1}     & $p^{j1}_{T}$      & $\eta_{j1}$  & $\phi_{j1}$  & $q_{j1}$  & -          & -        \\
\particle{j2}     & $p^{j2}_{T}$      & $\eta_{j2}$  & $\phi_{j2}$  & $q_{j2}$  & -          & -        \\
\particle{j3}     & $p^{j3}_{T}$      & $\eta_{j3}$  & $\phi_{j3}$  & $q_{j3}$  & -          & -        \\
\particle{b1}       & $p^{b1}_{T}$      & $\eta_{b1}$  & $\phi_{b1}$  & $q_{b1}$  & $m_{b1}$     & -        \\
\particle{b2}       & $p^{b2}_{T}$      & $\eta_{b2}$  & $\phi_{b2}$  & $q_{b2}$  & $m_{b2}$     & -        \\
\particle{lepton}   & $p^{l}_{T}$       & $\eta_{l}$   & $\phi_{l}$   & -         & -          & -        \\
\particle{energy}   & $E^{\text{Miss}}_{T}$    & -            & $\phi_{\text{ETMiss}}$ & -      & -        & $\sigma^{}_{\text{ETMiss}}$      \\
\hline
\end{tabular}
\end{table*}

\subsection{Mixture of Experts Graph Transformer}
\label{sec:mgt}
Given a graph $G$, representing a collision event as explained in Section \ref{sec:dataset}, our task is to predict a label $y$ such that $y = f(G)$. 
Specifically, the label $y$ encodes the class of the collision event that we aim to predict -- \textit{Signal} or \textit{Background} -- using a neural network model $f$. 
To ensure that the predictions are as interpretable as possible, we design the model $f$ with intrinsic explainability in mind and model it as a GT architecture \cite{GraphTransformer} enhanced with a MoE. The GT processes graph-structured data using self-attention and feed-forward networks, while the MoE enables parameter subsets to specialize in distinct data aspects. This architecture ensures intrinsic explainability by using attention to quantify node relationships. \Cref{fig:Transformer} illustrates the model, highlighting attention maps and the MoE’s role in capturing dependencies and specialization.

We test our proposed model on the Monte Carlo simulated collisions events for signal and SM background events described above. In this setup, each graph event $G$ is characterized by a feature matrix \( X \in \mathbb{R}^{N \times 6} \), where $N$ is the number of nodes in the graph and each row vector $x_i$ represents a node embedding with its features.
The GT updates nodes representations using a sequence of layers consisting in a self-attention mechanism followed by a feed-forward block. 
After the initial addition of Laplacian positional encodings \cite{LaplacianPE0}\cite{LaplacianPE}, \cite{LaplacianPE2}, for each node $x_i$, the attention mechanism computes a weighted sum of value vectors \( V \), where the weights are determined by scaled dot-product attention \cite{GraphTransformer}. Let \( Q, K, V \) denote the query, key, and value matrices derived from the input \( X \) via learned linear projections \( W_Q, W_K, W_V \). The attention for \( x_i \) is:
\begin{equation}
\text{Attention}(x_i) = \text{softmax}\left(\frac{q_i K^\top}{\sqrt{d_k}}\right) V
\end{equation} 
where \( q_i = x_i W_Q \) is the query vector for \( x_i \), \( K = X W_K \) is the matrix of key vectors, \( V = X W_V \) is the matrix of value vectors, and \( d_k \) is the dimensionality of the key vectors used for scaling.
We refer to $\text{softmax}\left(\frac{q_i K^\top}{\sqrt{d_k}}\right)$ as "attention weights", as they serve to encode the relationships between nodes, effectively quantifying their interactions and dependencies. This view offers an interpretable framework for understanding how the model processes information and makes decisions, which we describe more in detail in Section \ref{sec:attnmaps}. 
The GT employs multi-head attention, extending the attention mechanism by employing $H$ independent attention heads, each with its own set of projections \( W_Q^h, W_K^h, W_V^h \) for \( h = 1, \dots, H \). The outputs of these heads are concatenated and linearly projected back to the original dimensionality. For \( x_i \), the multi-head attention is:
\begin{equation}\label{attn}
\text{MultiHead}(x_i) = \text{Concat}(\text{head}_1, \dots, \text{head}_H) W_O, \qquad \text{with} \qquad \text{head}_h = \text{softmax}\left(\frac{q_i^h (K^h)^\top}{\sqrt{d_k}}\right) V^h
\end{equation}
where \( q_i^h = x_i W_Q^h \), \( K^h = X W_K^h \), and \( V^h = X W_V^h \). The output projection \( W_O \) combines the \( H \) heads into a single representation.
%
%
%
%
%
%
%
After updating the node representations according to \cref{attn}, the node embeddings are passed through a feed-forward network (FFN), consisting in a sequence (usually two) learnable linear projections with an intermediate nonlinearity:
\begin{equation}
x_i \mapsto W_2(\text{ReLU}(W_1(x_i) + \beta_1))+\beta_2
\label{eq:ffn}
\end{equation}
where $x_i$ represents a generic node embedding, $W_1, \beta_1, W_2, b_2$ represent learnable weights and biases, and $ReLU$ is a nonlinear activation function.

\begin{figure}[!ht]
    \centering
    \includegraphics[width=15.0cm]{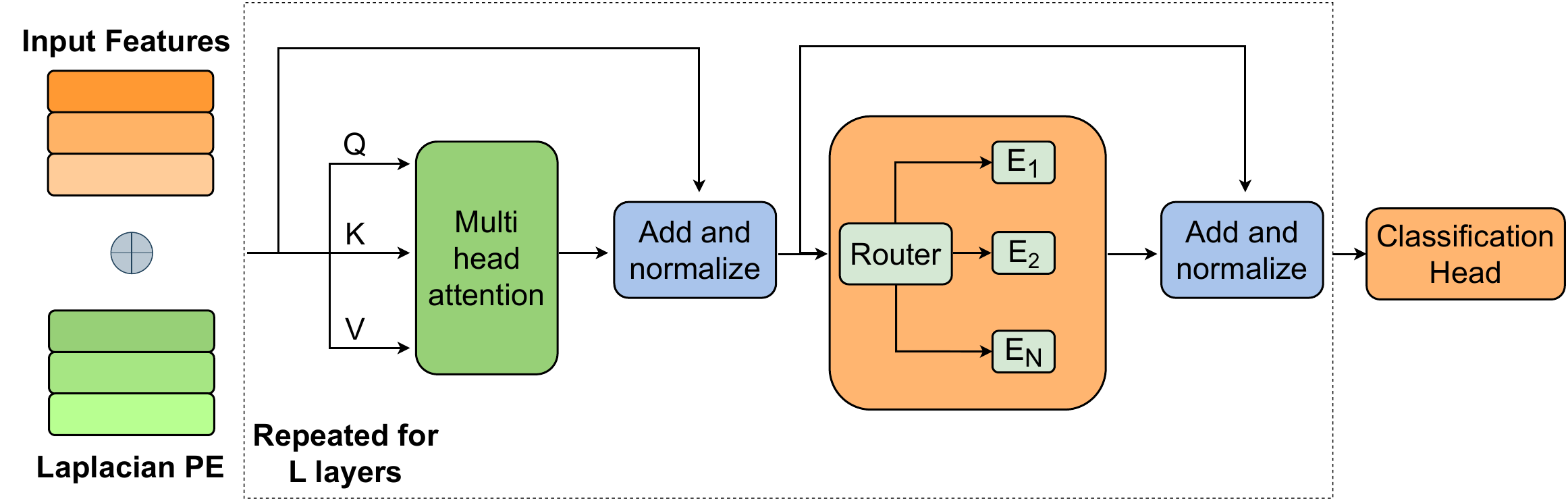}
    \caption{Detailed illustration of the proposed architecture incorporating multi-head attention with an MoE block. The model begins with Laplacian positional encoding for the input features, followed by multi-head attention and normalization. This structure is repeated across L layers. The MoE block, driven by a gating network, assigns inputs dynamically to specialized expert networks. A classification head processes the final representation to produce predictions.}
    \label{fig:MoE}
\end{figure}

In order to enhance explainability, we also replace the original FFN with a MoE block, as described next.
The MoE architecture consists of $n$ expert networks, denoted as $E_1, \dots, E_n$, and a routing network $r$. Each expert is a neural network defined as in Eq. \eqref{eq:ffn} with its own parameters. 
The routing network $r$ determines a set of weights for the expert outputs based on the updated node representation, thus assigning each node of the graph to one or more expert networks.
The output $y$ of the MoE block is computed as a weighted combination of the expert outputs:
\begin{equation*}
y = \sum_{i=1}^n r({x})_i E_i({x}).
\end{equation*}
The MoE enhances explainability by allowing us to identify which experts contribute most to the prediction for a given node, providing insights into the model's decision-making process.
Additionally, the MoE optimizes computational efficiency by leveraging the sparsity of the gating network's output. When $r(x)_i = 0$, we bypass the computation of $E_i(x)$, thus saving computational resources. 
We follow (Shazeer et al, $2017$)\cite{MoE} for the design of the routing network, and implement noisy \textit{top-K} gating with a load balancing loss to prevent imbalanced routing choices that would lead to the training of a small subset of experts, a well-known issue for MoE models. Details about the routing are given in S1 of Supplementary Materials.
The overall architecture is further depicted in  \cref{fig:MoE} and includes the GT stack and the classification head tailored to the task.

\section{Results}
We conducted experiments to evaluate the performance of four architectures—Graph Convolutional Network (GCN), MLP, GT, and our model described in \cref{sec:mgt}, the Mixture of Experts Graph Transformer (MGT)—on the SUSY dataset. The GCN architecture, uses two convolutional layers with hidden channels of size 80 and 40, respectively. It employs global mean pooling for aggregation and concludes with a linear layer to output the class predictions. The GCN results show promising differentiation between signal and background, aligning with key kinematic features of the dataset, with detailed analysis, including feature importance and event correlations, provided in S4 of Supplementary Materials. The MLP model also consists of two hidden layers, with dimensions 80 and 40, respectively, and takes the input graph as a flattened vector. The GT architecture is composed of two layers, each with 2 attention heads, Laplacian positional embeddings of size 4 and a hidden size of 80. The MGT model builds upon the GT architecture with a hidden size of 80, 2 attention heads, 6 expert networks each with an expert size of 20, and 2 layers. It also employs a gating mechanism with a weight load factor of 1 for expert selection. 

The experiments were conducted using ten runs, with a different random seed for each run, to ensure robust evaluation of model performance. The results, quantified in terms of accuracy, precision, recall, F1-score, and AUC are reported in \cref{tab:performance_comparison}. The MoE architecture outperforms the other models, achieving superior accuracy and overall performance metrics. This highlights the effectiveness of MoE in leveraging expert networks for enhanced predictive capabilities. Further analysis on training is provided in S3 of the Supplementary Materials.

\begin{table*}[ht]
\centering
\caption{Performance comparison of different architectures (MLP, GCN, GT, and MGT) on the SUSY dataset. Results are reported as the mean $\pm$ standard deviation over ten runs with different random seeds. Metrics include Accuracy, Precision, Recall, F1 score, and AUC.}
\label{tab:performance_comparison}
\begin{tabular}{lcccccc} 
\hline
Model & Accuracy & Precision & Recall & F1 & AUC  \\ 
\hline
GCN & 0.750 $\pm$ 0.0022 & 0.779 $\pm$ 0.2549 & 0.700 $\pm$ 0.0427 & 0.736 $\pm$ 0.0131 & 0.832 $\pm$ 0.0134 \\
MLP & 0.829 $\pm$ 0.0015 & 0.826 $\pm$ 0.0054 & 0.835 $\pm$ 0.0101 & 0.830 $\pm$ 0.0027 & 0.913 $\pm$ 0.0017 \\
GT & 0.849 $\pm$ 0.0059 & 0.850 $\pm$ 0.0062 & 0.849 $\pm$ 0.0059 & 0.849 $\pm$ 0.0060 & 0.928 $\pm$ 0.0057
 \\
MGT & $\bm{0.852}$ $\pm$ $\bm{0.0005}$ & $\bm{0.851}$ $\pm$ $\bm{0.0008}$ & $\bm{0.849}$ $\pm$ $\bm{0.0016}$ & $\bm{0.854}$ $\pm$ $\bm{0.0004}$ & $\bm{0.929}$ $\pm$ $\bm{0.0039}$  \\
\hline
\end{tabular}
\end{table*}

\section{Discussion}
\begin{figure}[!ht]
    \centering
    \includegraphics[width=15.0cm]{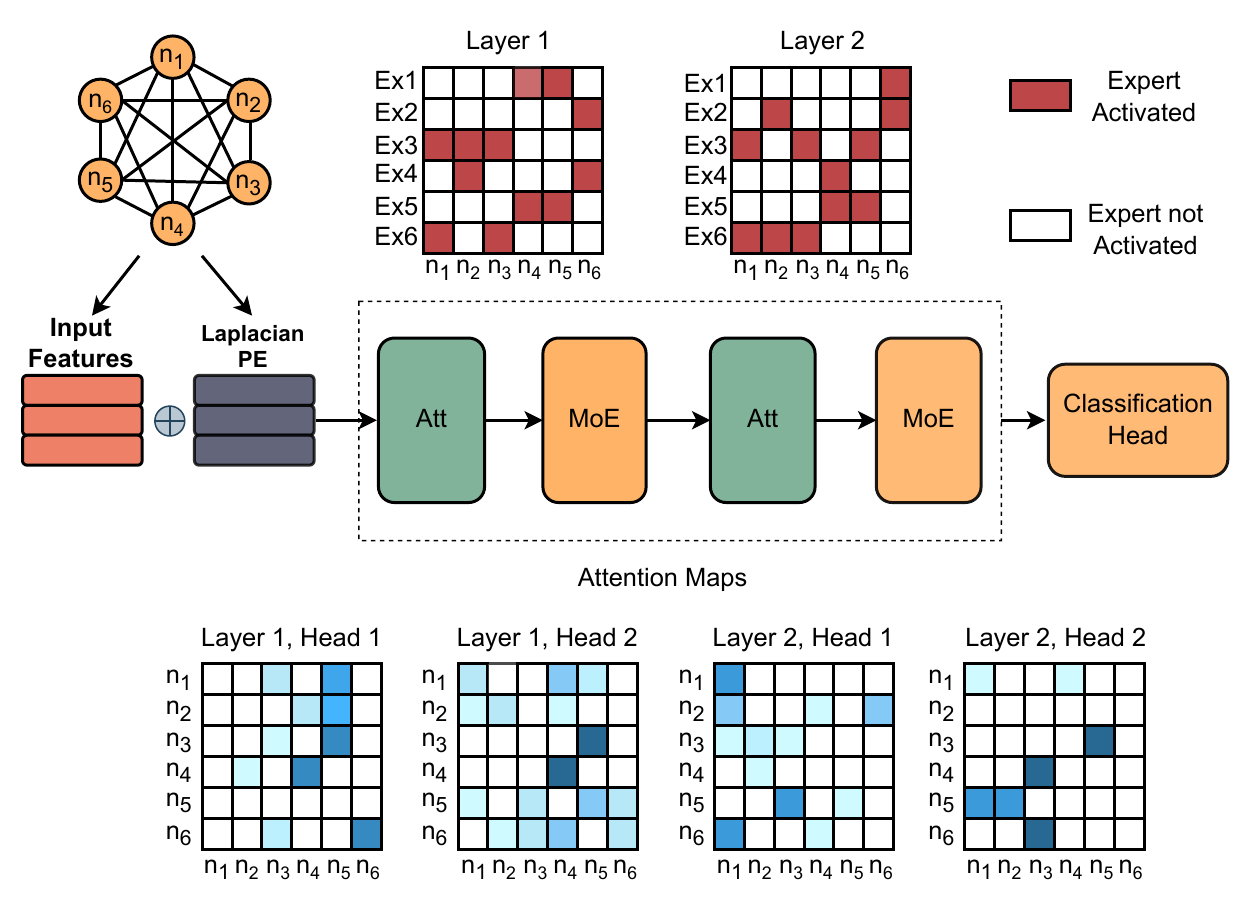}
    \caption{Overview of the proposed Transformer-based model architecture: the image illustrates an example of the model, which takes as input a graph. The graph is processed by the model, comprising various blocks: \textbf{Att}, which is the Multi-Head Attention block, and \textbf{MoE} blocks. The visualization includes attention maps derived from the Multi-Head Attention mechanism and the activation patterns of the experts for a single collision event example.}
    \label{fig:Transformer}
\end{figure}
The attention mechanisms and expert specialization in the MoE model provide valuable insights into its explainability and performance. This section discusses the detailed behavior of attention scores and the unique specialization patterns exhibited by expert networks. The overall architecture of the proposed Transformer-based model is illustrated in Figure~\ref{fig:Transformer}, providing a detailed view of how the graph input is processed through the MoE blocks.
\subsection{Analysis of the Attention Heads}
\label{sec:attnmaps}
\begin{figure}[!ht]
    \centering
    \begin{subfigure}{0.49\linewidth}
        \centering
        \includegraphics[width=\linewidth]{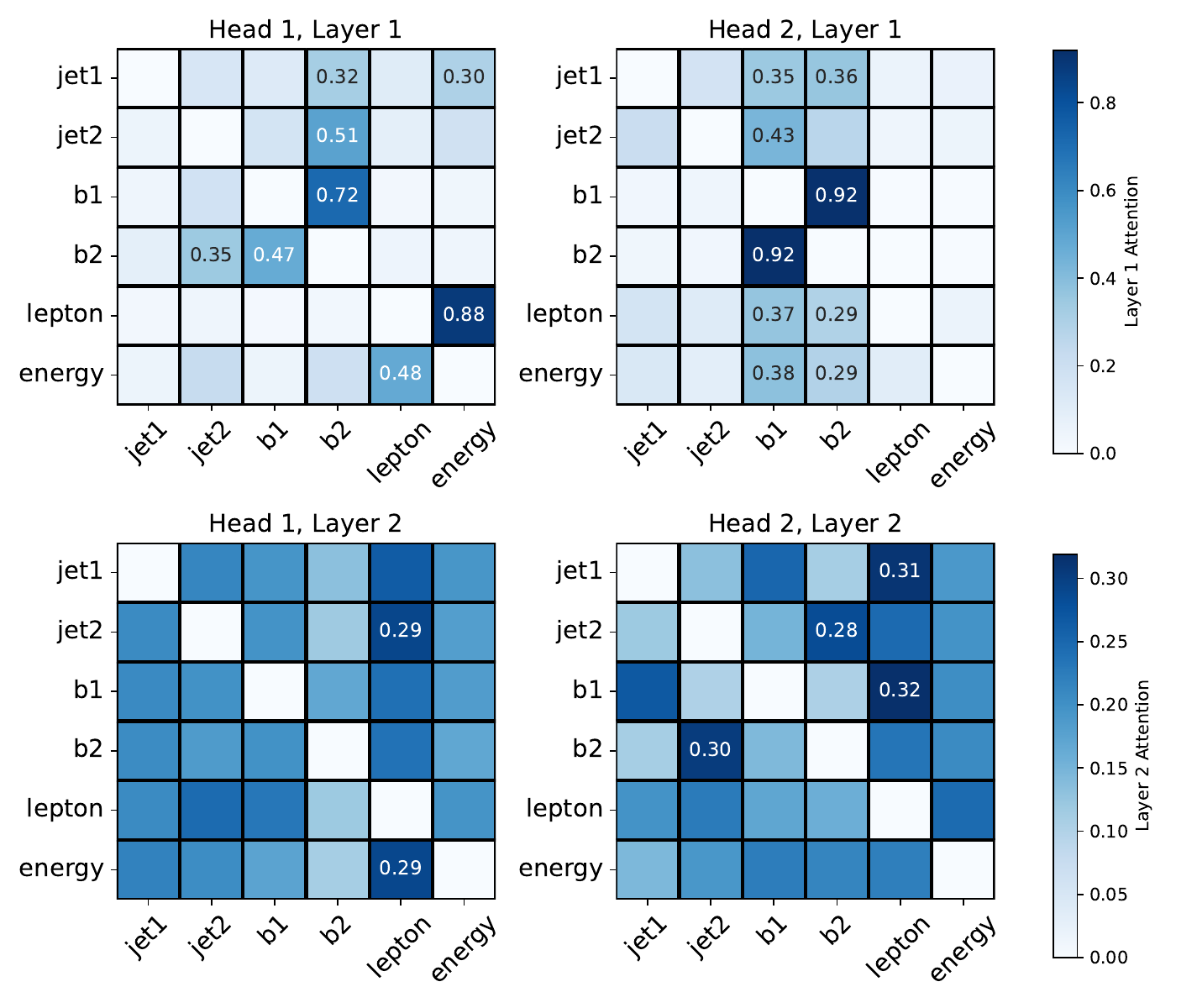}
        \caption{6 nodes.}
        \label{fig:attention_heatmap_test_6}
    \end{subfigure}
    \hfill
    \begin{subfigure}{0.49\linewidth}
        \centering
        \includegraphics[width=\linewidth]{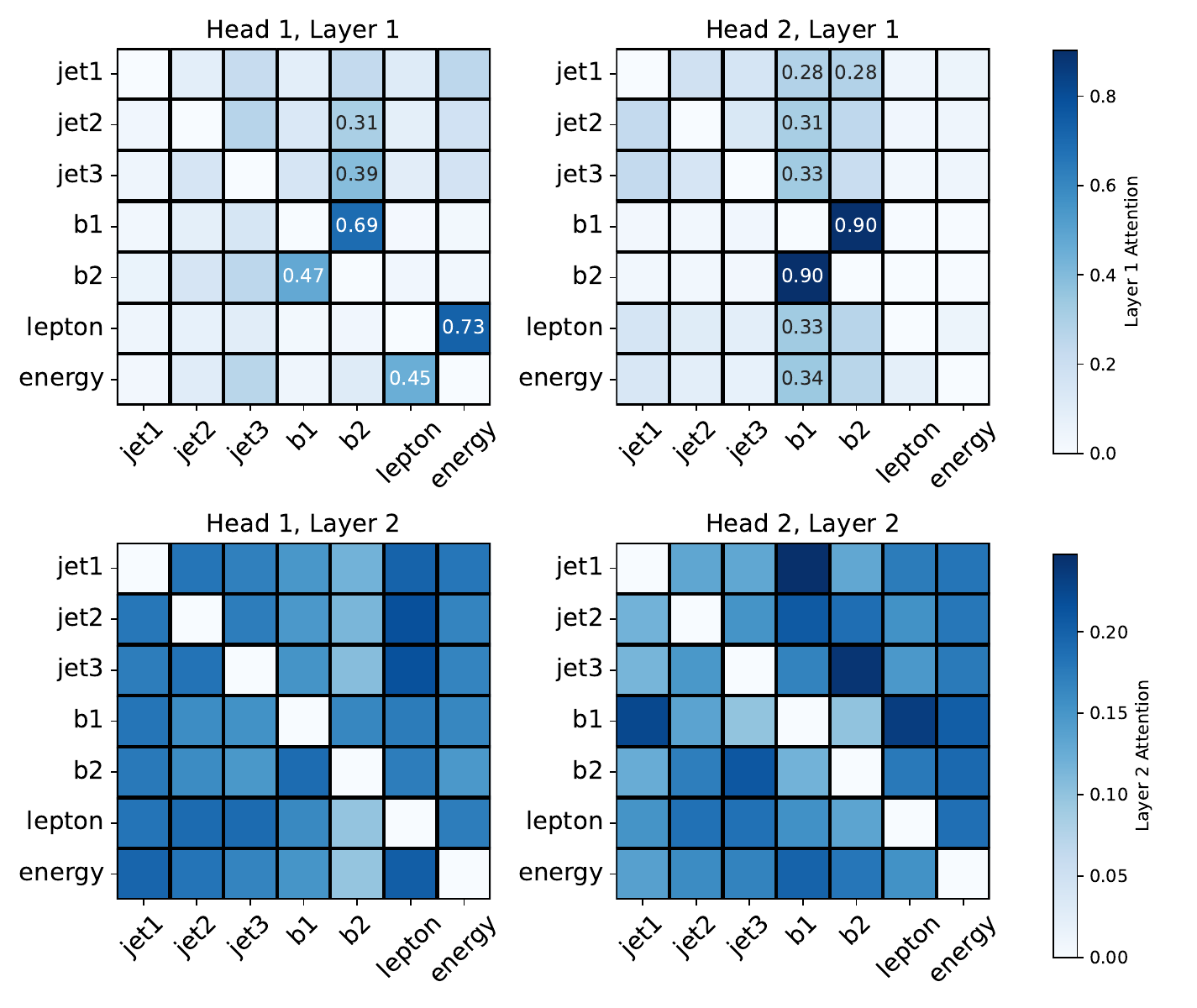}
        \caption{7 nodes.}
        \label{fig:attention_heatmap_test_7}
    \end{subfigure}
    \caption{Attention maps for test set.Each image shows the averaged attention scores for a specific attention head. The color scale ranges from white (low attention) to dark blue (high attention), visualizing only attention scores above the 85th percentile (computed across all layers and heads) to highlight the most significant interactions. The left column displays maps for 6-node graphs; the right column for 7-node graphs.}
    \label{fig:attention_heatmap_test}
\end{figure}

\begin{figure}[!ht]
    \centering
    \begin{subfigure}{0.49\linewidth}
        \centering
        \includegraphics[width=\linewidth]{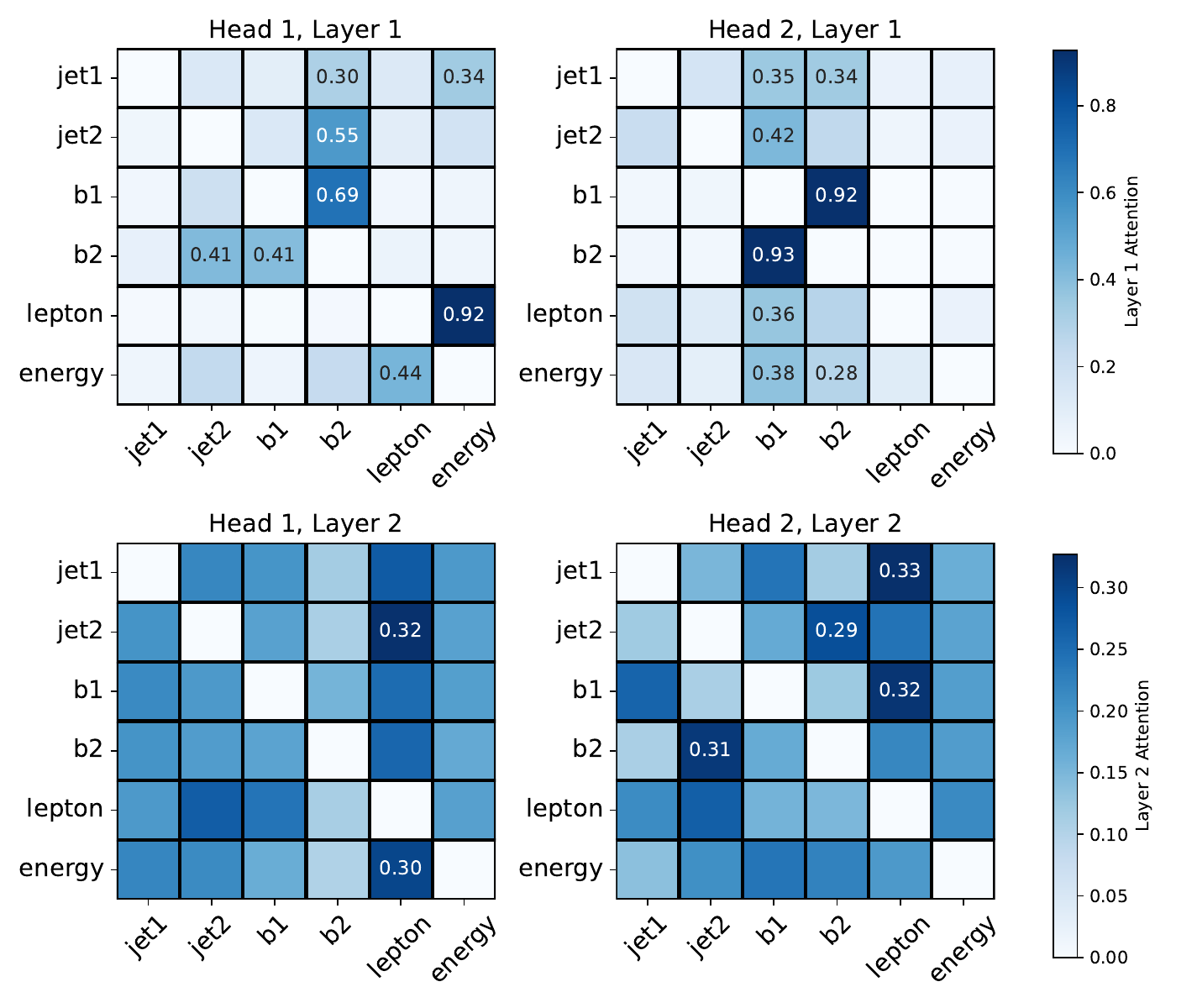}
        \caption{6 nodes.}
        \label{fig:attention_heatmap_corsign_6}
    \end{subfigure}
    \hfill
    \begin{subfigure}{0.49\linewidth}
        \centering
        \includegraphics[width=\linewidth]{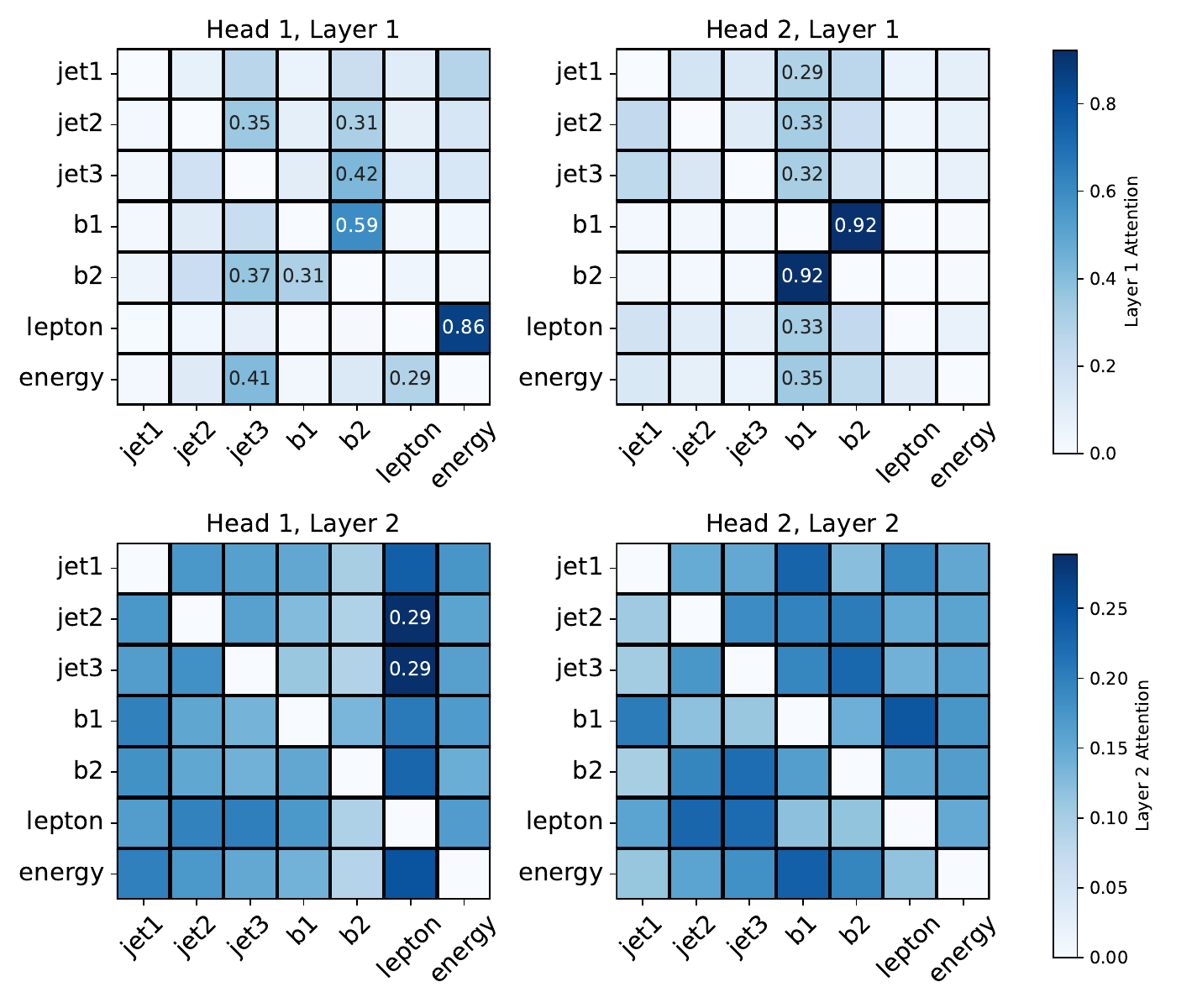}
        \caption{7 nodes.}
        \label{fig:attention_heatmap_corsign_7}
    \end{subfigure}
    \caption{Attention maps for correclty classified signals (true positives). Each image shows the averaged attention scores for a specific attention head. The color scale ranges from white (low attention) to dark blue (high attention), visualizing only attention scores above the 85th percentile (computed across all layers and heads) to highlight the most significant interactions. The left column displays maps for 6-node graphs; the right column for 7-node graphs.}
    \label{fig:attention_heatmap_corsig}
\end{figure}

\begin{figure}[!ht]
    \centering
    \begin{subfigure}{0.49\linewidth}
        \centering
        \includegraphics[width=\linewidth]{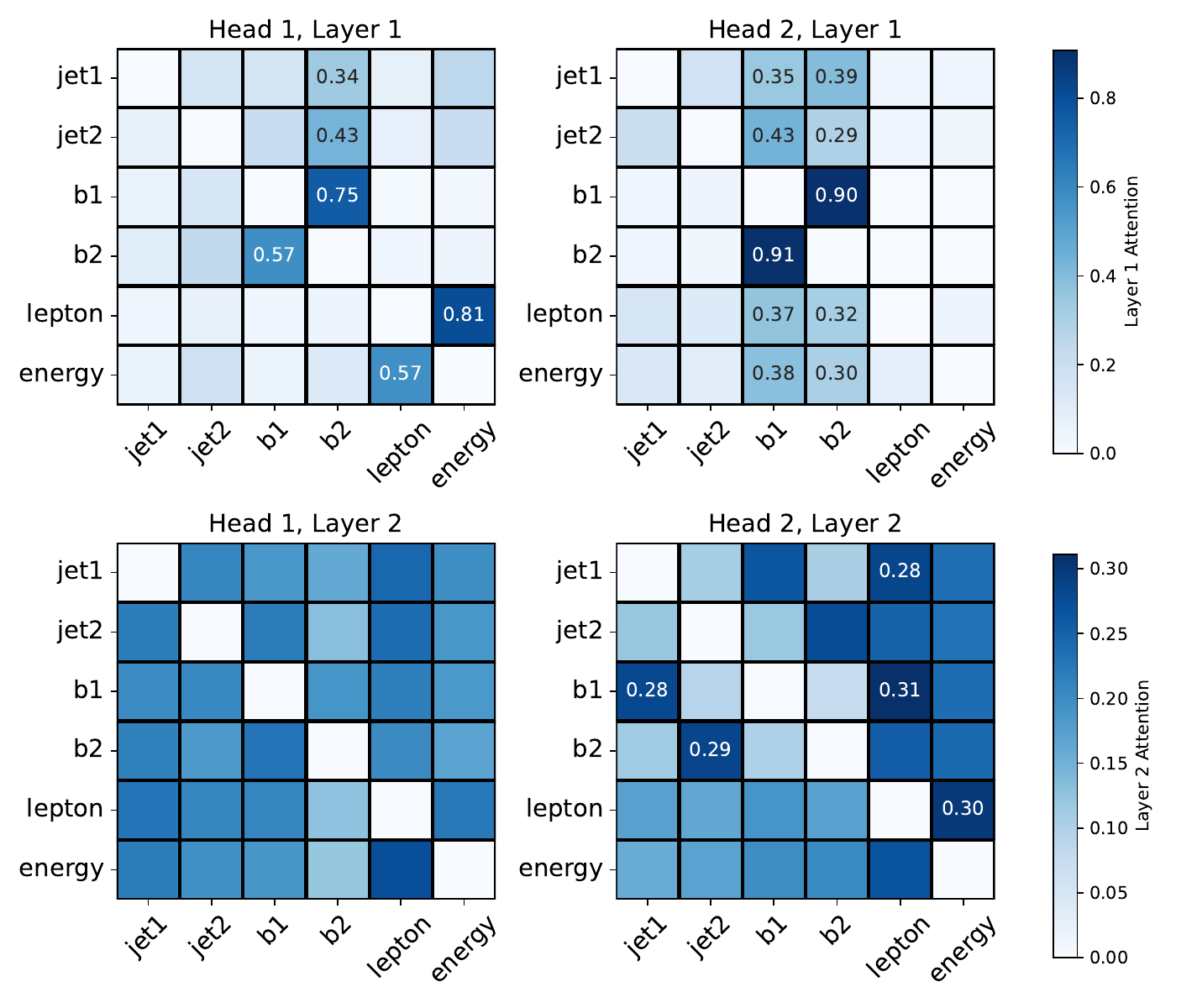}
        \caption{6 nodes.}
        \label{fig:attention_heatmap_corback_6}
    \end{subfigure}
    \hfill
    \begin{subfigure}{0.49\linewidth}
        \centering
        \includegraphics[width=\linewidth]{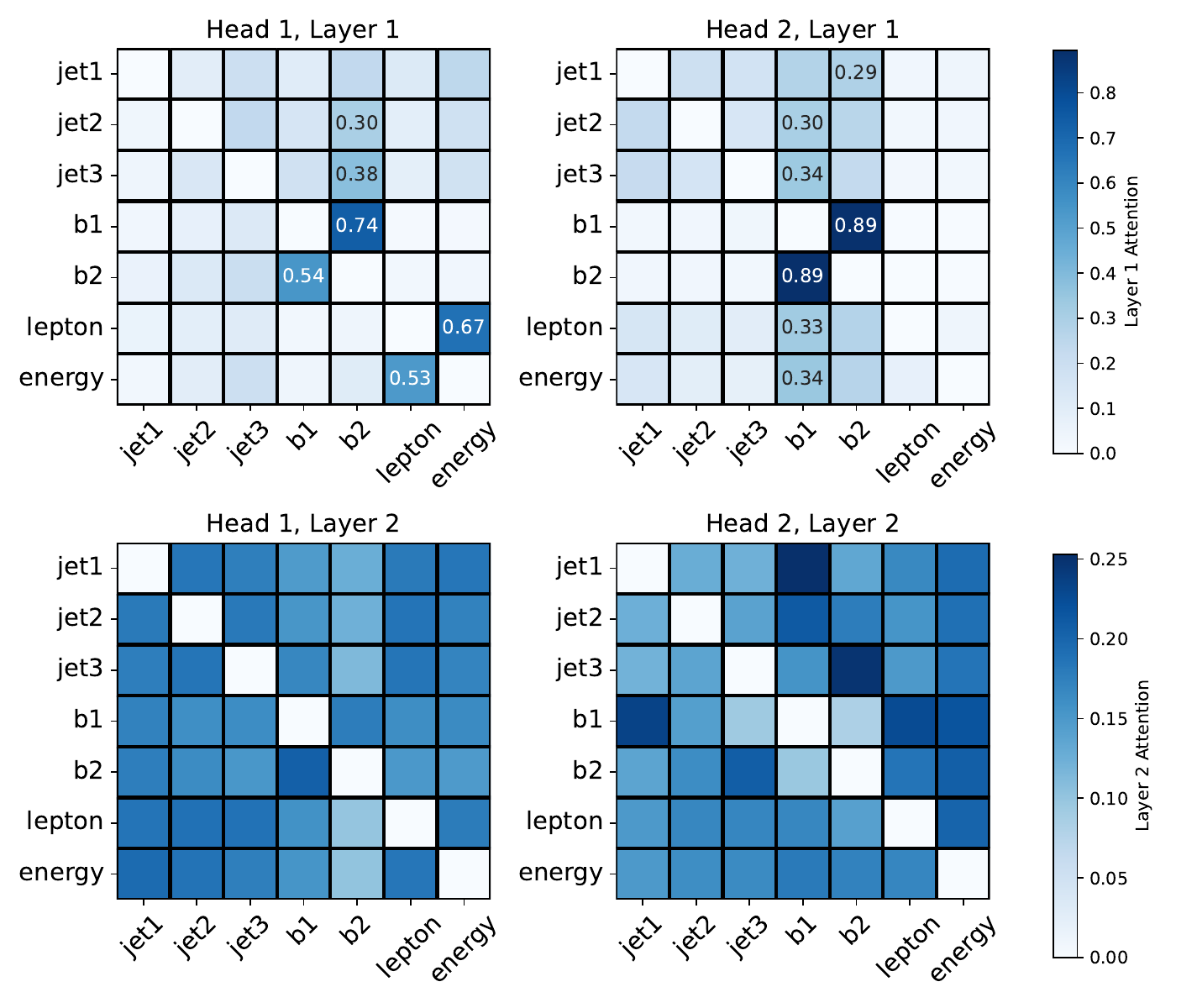}
        \caption{7 nodes.}
        \label{fig:attention_heatmap_corback_7}
    \end{subfigure}
    \caption{Attention maps for correctly classified backgrounds (true negatives). Each image shows the averaged attention scores for a specific attention head. The color scale ranges from white (low attention) to dark blue (high attention), visualizing only attention scores above the 85th percentile (computed across all layers and heads) to highlight the most significant interactions. The left column displays maps for 6-node graphs; the right column for 7-node graphs.}
    \label{fig:attention_heatmap_corback}
\end{figure}

Attention scores provide significant insights into the model's internal mechanisms by refining node and feature representations while capturing relationships and dependencies within the graph. In this process, attention acts as a guiding mechanism, directing how information is aggregated and propagated through the network.  In an attention map, the attention weights are organized such that each row corresponds to a query (i.e., the element doing the attending), and each column corresponds to a key (i.e., the element being attended to). 
To interpret the model, we conduct a detailed analysis of the attention scores by examining multiple subsets of the dataset, including the training set, test set, as well as correctly classified signal events (true positives) and correctly classified background events (true negatives). For each of these groups, we compute the attention matrix for each head, averaging the values across the data to uncover global trends in attention behavior. The images shown in the following (\cref{fig:attention_heatmap_test}, \cref{fig:attention_heatmap_corsig}, \cref{fig:attention_heatmap_corback}) are divided into two columns, with 6-node graphs on the left and 7-node graphs on the right. Each row represents the attention given by a specific node to all other nodes in the graph. Attention scores above the 85th percentile, computed across all layers and heads, are visualised to highlight the most significant interactions while reducing noise. By visualizing the attention maps, we can diagnose how attention guides the formation of feature representations, shedding light on the underlying mechanisms of the model. 
\cref{fig:attention_heatmap_test} shows the resulting image for the test set. Attention in the first layer displays higher magnitudes and broader distributions, capturing diverse relationships within the graph. In contrast, the second layer shows reduced magnitudes, indicating a refinement of features. This hierarchical approach reflects the model's process of first aggregating general dependencies and then fine-tuning these into task-relevant representations. 
The attention focus in head 2 is dominated by core features of \particle{b1}, \particle{b2}, whereas in head 1 the \particle{energy}, \particle{lepton} plays a more central role, with \particle{b1}, \particle{b2} remaining of secondary relevance. This behaviour is consistently observed across both graph types. 
This is in excellent agreement with the physics expectations: the origin of the \particle{b1},\particle{b2} is a Higgs boson in signal that is not present in background, and as such is clearly distinctive; the missing transverse energy, in particular when related to the lepton object, is expected to be the next distinctive feature because it originates from the dark matter particles and the neutrino in signal, and only from the neutrino in background, with lepton and neutrino originating from the decay of the $W$ boson (both in signal and background).   

The above considerations are even better depicted when considering correctly classified signals (\cref{fig:attention_heatmap_corsig}), where attention strongly concentrates on \particle{b1}, \particle{b2}, and \particle{energy} nodes, highlighting their importance in distinguishing signal events. In contrast, for correctly classified backgrounds
(\cref{fig:attention_heatmap_corback}), attention is more evenly distributed across nodes, reflecting the broader and less distinctive feature patterns typical of background events. Finally, we note how the additional jet node in 7-node graphs receives some attention in the second layer for signal events, indicating a refinement of features based on its correlation with the other jets in the event. More specifically, the attention score of \particle{j3} in 7-node graphs can be explained by the fact that the feature $\eta$ is expected to have some distinctive power, being in average lower for background and larger for signal. The quantile of such extra jet is also an additional distinctive feature. 

\subsection{Analysis of Experts Specialization}
\begin{figure}[!ht]
    \centering
    \includegraphics[width=12.0cm]{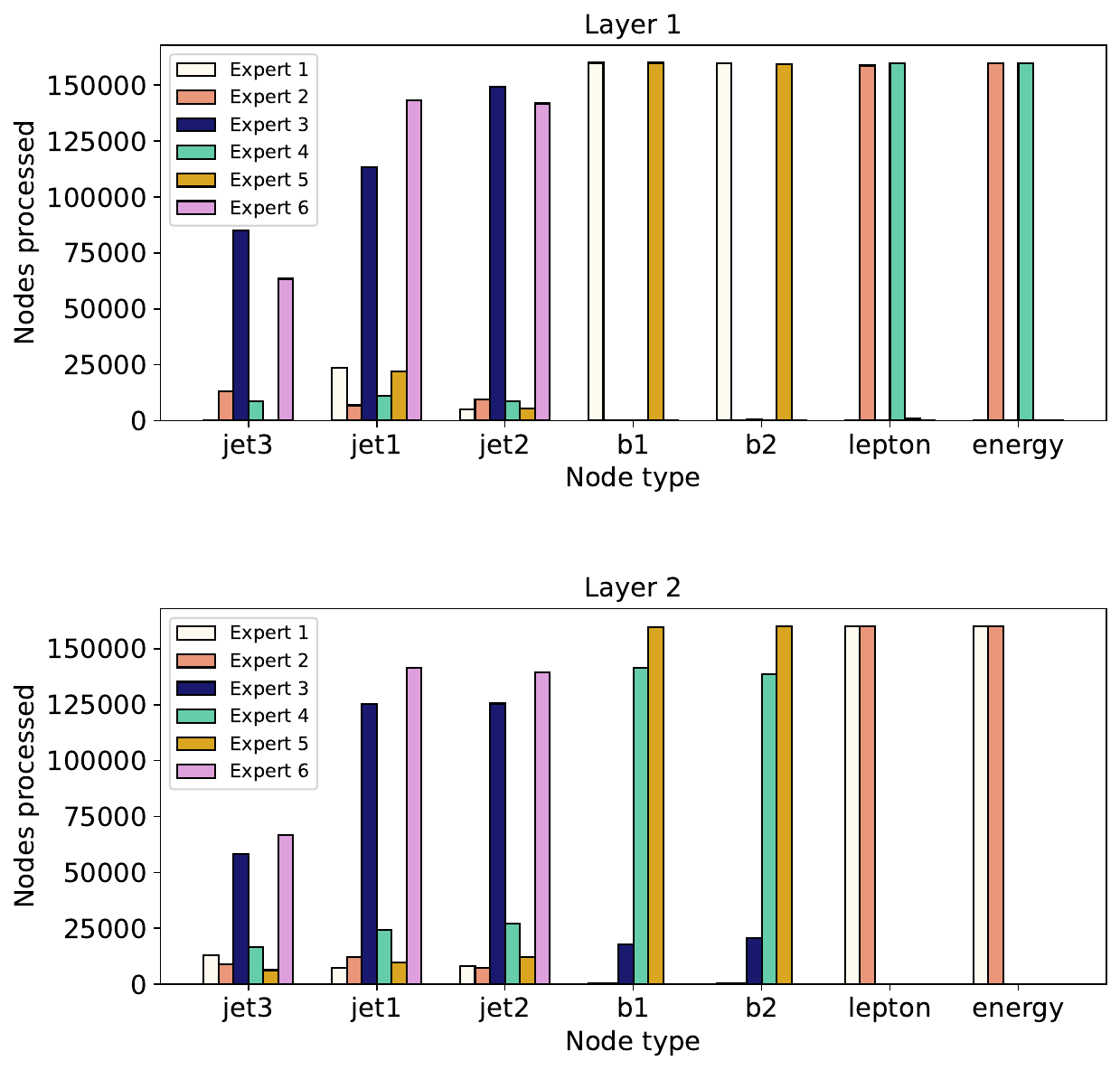}
    \caption{Visualization of node specialization in the MoE architecture for layers 1 and 2. The bar charts show the number of nodes processed by each expert for different node types (\particle{jet1}, \particle{jet2}, \particle{jet3}, \particle{b1}, \particle{b2}, \particle{lepton}, \particle{energy}).}
    \label{fig:Expert Specialization}
\end{figure}

Unlike traditional neural network approaches that rely on a single monolithic network, MoE introduces a dynamic mechanism which can be helpful in processing and interpreting complex data. 
The key motivation behind employing MoE in our GT model is to understand how different expert networks specialize in processing distinct aspects of the interaction graph. 
We test node specialization which specifies how the model processes inputs by dividing the workload among its components. Each specialized expert becomes highly tuned to a certain type of input (in our case the type of node), allowing it to extract and represent the most relevant information from that subset.
Using the same dataset of signal and background events, the specialization patterns of the MoE block for layers 1 and 2 is studied. \cref{fig:Expert Specialization}) highlights the distinct ways the model processes the features of different nodes. The MoE block of layer 1 appears to focus on extracting fundamental distinctions between node types, with strong specialization and minimal overlap between experts.
The MoE block of layer 2, on the other hand, shows slightly more overlap among experts, suggesting that it operates at a higher level of abstraction, combining features and resolving correlations between particle types to refine the classification. 
In Tables \ref{tab:expert_tab_1} and \ref{tab:expert_tab_2}, we present a summary of the specializations of the various experts, highlighting the nodes where the experts exhibit the highest activation.
In more detail, for the MoE block of layer 1, Experts 2 and 4 focus on \particle{lepton} and \particle{energy} nodes, which are the quantities closely tied to the W boson. As highlighted before, in background events, the missing energy arises solely from the neutrino produced in the decay of the W boson whilst in signal events missing energy also originates from dark matter candidates. The specialization of Experts 2 and 4 indicates that the model identifies the combined behavior of \particle{lepton} and missing energy features as a key discriminator. Together, these quantities encode differences between signal and background events, such as the overall energy imbalance or the transverse momentum carried away by undetected particles. \
The paired specialization of Experts 1 and 5 on the two  
b-jets suggests that the model learns to process b-jets as a correlated entity rather than treating them in isolation. This approach is consistent with the physics. 
Experts 3 and 6 demonstrate specialization in \particle{jet1}, \particle{jet2}, and to a lesser degree, \particle{jet3}. This pattern can be explained by the structure of the dataset. For 6-node graphs, \particle{jet1}, \particle{jet2} are equivalent to \particle{b1}, \particle{b2}, hence the same specialization of Experts 1 and 5 is expected. For 7-node graphs, \particle{jet3} acquires an added distinctive value. The fact that it is not consistently present across all events limits the opportunities for these experts to learn from its features.    
The MoE block of layer 2, on the other hand, shows slightly more overlap among experts. This may suggest that layer 2 operates at a higher level of abstraction—combining features and resolving correlations between particle types to refine the classification—the qualitative nature of this observation does not allow us to make a definitive quantitative claim.

\color{black}

\begin{table*}[!ht]
\centering
\caption{Expert Specialization for layer 1.}
\label{tab:expert_tab_1}
\begin{tabular}{>{\centering\arraybackslash}m{2cm}lcc} 
\hline
Expert Number & Specialization & Physical Explanation  \\ 
\hline
1, 5 & \particle{b1}, \particle{b2} & \makecell{Experts focus on the b-jets as a correlated entity. \\ In signal events, these arise from Higgs boson decay, \\ but in background events, they do not.} \\ \\
2, 4 & \particle{lepton}, \particle{energy} & \makecell{Experts specialize in detecting lepton and missing energy features, \\ key indicators for distinguishing signal from background. \\ In signal, missing energy also originates from dark matter particles, \\ unlike the background.}  \\ \\
3, 6 & \particle{jet1}, \particle{jet2}, \particle{jet3} & \makecell{Equivalent to \particle{b1} and \particle{b2} in 6-node graphs, \\ but in 7-node graphs,  the additional jet (\particle{jet3}) introduces \\ distinct kinematic features that alter the specialization} \\
\hline
\end{tabular}
\end{table*}


\begin{table*}[!ht]
\centering
\caption{Expert Specialization for layer 2.}
\label{tab:expert_tab_2}
\begin{tabular}{>{\centering\arraybackslash}m{2cm}lcc} 
\hline
Expert Number & Specialization & Physical Explanation  \\ 
\hline
1, 2 & \particle{lepton}, \particle{energy} & Similar to Experts 2 and 4 in Layer 1 \\
3, 6 & \particle{jet1}, \particle{jet2}, \particle{jet3} & Similar to Experts 3 and 6 in Layer 1, but with less specialization \\
4, 5 & \particle{b1}, \particle{b2} & Similar to Experts 1 and 5 in Layer 1, but with less specialization
 \\
\hline
\end{tabular}
\end{table*}

\section{Conclusion}
In this work, we introduced a novel Mixture of Experts layer for graph data, aiming to enhance both explainability and performance in high-energy particle physics analysis. The proposed model, evaluated on a HEP dataset of simulated events from a search for dark matter in SUSY models published by the ATLAS experiment, achieved superior predictive accuracy compared to baseline models. This improvement underscores the effectiveness of leveraging expert subnetworks to address the task-specific complexities inherent in high-energy physics data.
Beyond its strong predictive performance, the model made significant strides in addressing the critical need for explainability in high-energy physics. By visualizing attention maps and analyzing expert specializations, we identified key features and their influence on classification outcomes, ensuring that the model’s decision-making aligns with established physical principles. This ability to trace predictions back to interpretable components not only enhances trust in the model but also provides a deeper understanding of the physics involved.
Despite these achievements, the study has some limitations that warrant further exploration. The scalability of the MGT model to larger datasets or graphs with more nodes remains an open question, as does its ability to generalize across a wider range of high-energy physics tasks. Moreover, future validations of these results on other datasets and larger architectures will be critical in establishing their robustness and broader applicability.
Future research could extend this framework to other types of datasets and particle physics phenomena, testing its adaptability and robustness across diverse scenarios. Furthermore, these results are particularly promising in relation to the application of advanced mechanistic explainability techniques, such as  Sparse Autoencoders \cite{SAE}, to this type of data. The integration of emerging techniques for automated explainability, including the use of Large Language Models \cite{LLMXAI}, could further enhance the accessibility and dynamism of the model's interpretations, amplifying its utility for scientific discovery. 
By combining high predictive performance with enhanced explainability, this work lays the groundwork for advancing explainable machine-learning methodologies in particle physics. Building on these foundations promises to unlock further potential in addressing the challenges of analyzing complex scientific data.

\section*{Data Availability Statement}
The data used is public and released under Creative Commons Zero v1.0 Universal license. It is fully accessible via opendata \href{https://opendata.cern.ch/record/28100}{https://opendata.cern.ch/record/28100}, where a description of the content is included. The generation and simulation setup are fully detailed in the cited ATLAS paper (\href{https://link.springer.com/article/10.1007/JHEP12(2023)167}{10.1007/JHEP12(2023)167}). The code for the model is available at \href{https://github.com/DonatellaGenovese/Mixture-of-Expert-Graph-Transformer/tree/main}{https://github.com/DonatellaGenovese/Mixture-of-Expert-Graph-Transformer}.

\newpage
\section*{Supplementary materials}
\subsection*{S1 Mixture of Experts Routing and Balancing loss}
\label{S1}

The Mixture of Experts (MoE) architecture provides a more efficient way to allocate resources by specializing subnetworks (Experts), enabling the possibility for sparse computing. This design is a core component of the Mixture of Experts Graph Transformer (MGT) introduced in \cref{sec:mgt}, where it replaces the standard feed-forward networks in the Graph Transformer (GT) layers. By integrating MoE, the MGT achieves both computational efficiency and enhanced explainability. 

At the core of the MoE layer in the MGT is the concept of specialization, where multiple expert networks learn to focus on different regions of the input space. Instead of employing a monolithic model to process every input, MoE uses a gating mechanism that dynamically selects the most relevant experts for each node representation. This dynamic selection not only reduces the computational cost by utilizing a sparse subset of experts for each input but also enables detailed interpretability by identifying which experts were most influential in the decision-making process. 

In the MGT architecture (\cref{sec:mgt}), the MoE layer is applied to updated node embeddings after the attention mechanism. The routing network, as detailed below, determines the top-k experts for each node and provides the weights that combine their outputs. The gating mechanism employs Noisy Top-K Gating \cite{MoE}, and its implementation aligns closely with the MGT design goals outlined in the main text.
The gating function used in the MoE layer is:
\begin{equation*}
    G(\hat{x}^{l})= \text{Softmax}(\text{KeepTopK}(H(\hat{x}^{l}),k))
\end{equation*}
where
\begin{equation*}
    H(\hat{x}^{l})_i= (\hat{x}^{l} \cdot W_g)_{i} + \text{Standardnormal}() \cdot \text{softplus} (\hat{x}^{l} \cdot W_{noise})_{i}
\end{equation*}
Here, \( W_g \) is the router matrix,$StandardNormal()$ is a random vector of dimensions $|E|$ with values drawn from a standard normal distribution, \( W_{noise} \) is a trainable weight matrix and the function 
\(\text{KeepTopK}(v, k)_i\) is defined as:
\begin{equation*}
    \text{KeepTopK}(v,k)_i= \begin{cases}
    v_i\ \ \ \ \text{if} \ v_i \ \text{is in the} \ k  \ \text{highest value components of} \ v \\
    - \infty \ \ \ \ \text{otherwise}
    \end{cases}  
\end{equation*}\\
Each expert is defined as follows:
\begin{equation}
E(x) = \text{Dropout}(W_2(\text{LeakyReLU}(W_1(x) + \beta_1))+\beta_2)
\label{eq:ex}
\end{equation}\\
\noindent The input node is routed to the experts corresponding to the $k$ non zero components of its gating function output.\\
The result of the expert's computation is combined as follows:
\begin{equation}
    y=\sum_{i=1}^k G(x)_i E_i(x)
\end{equation}\\

\noindent To ensure the even utilization of the experts, an additional loss component for load balancing is needed.\\
It is computed as follows:
\begin{equation}
L_{\text {load }}(X)=w_{\text {load }} \cdot \text{CV}(\operatorname{Load}(X))^2
\end{equation}
With $w_{load}$ being an hyperparameter factor that will control the weight of the load balancing loss with respect to the criterion, $CV$ being the coefficient of variation and $X$ being a batch of inputs.\\
The loads of the experts are computed as follows:
\begin{equation}
    \operatorname{Load}(X)_i=\sum_{x \in X} P(x, i) \\
\end{equation}\\
where
\begin{equation}
    P(x, i)=\Phi\left(\frac{\left(x \cdot W_g\right)_i-k t h \_ \text {excluding }(H(x), k, i)}{\operatorname{Softplus}\left(\left(x \cdot W_{\text {noise }}\right)_i\right)}\right)
\end{equation}
Here, $\Phi$ is the cumulative distribution function of the standard normal distribution, the term $kthexcluding(H(x),k,i)$ stands for the $k$th highest component of $H(x)$, excluding component $i$.\\

\begin{figure}
    \centering
    \includegraphics[width=12.0cm]{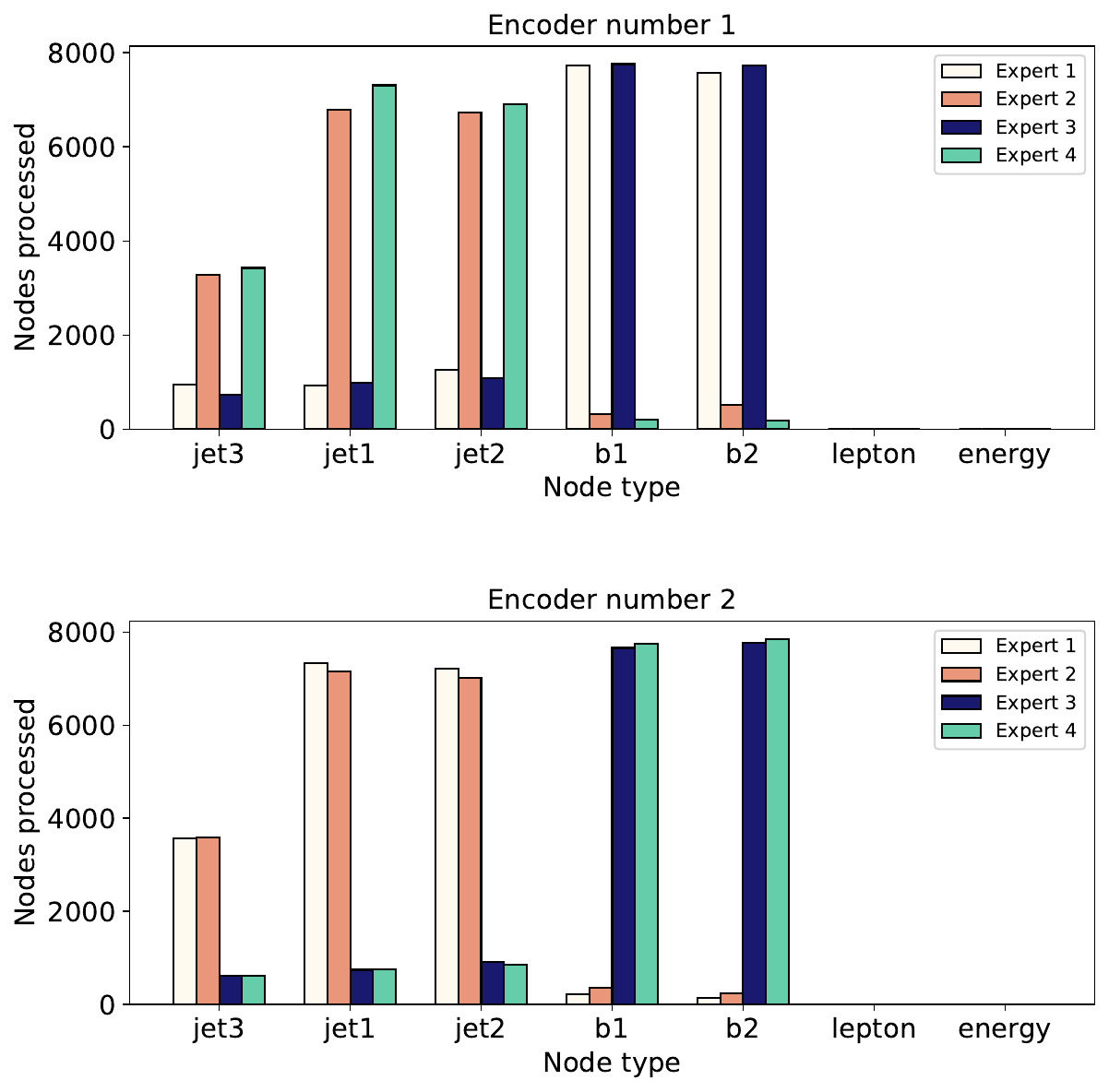}
    \caption{Expert specialization pattern with 4 total experts, top-2 routing, 2 layers}
    \label{fig:4ex}
\end{figure}
\begin{figure}
    \centering
    \includegraphics[width=12.0cm]{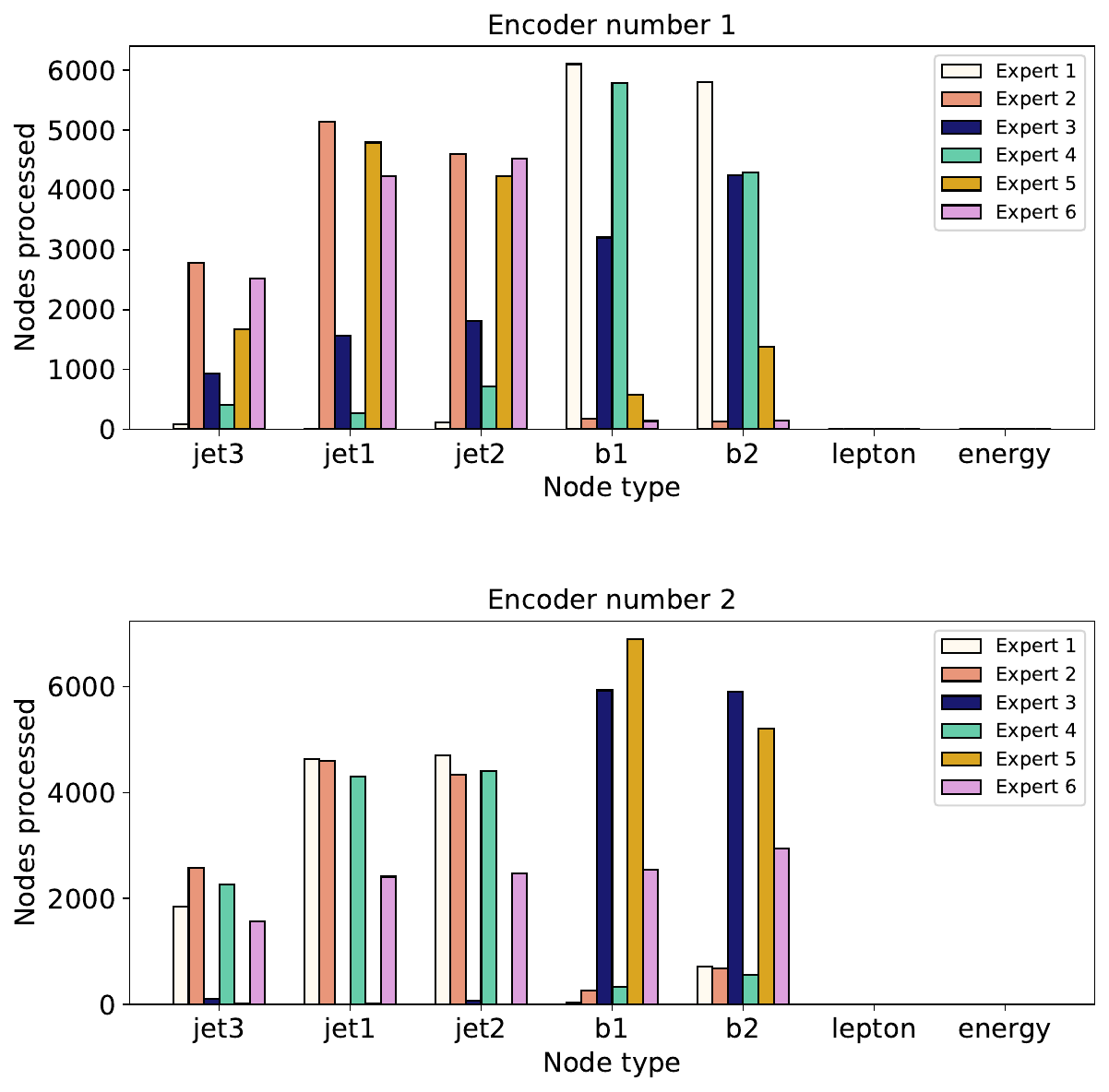}
    \caption{Expert specialization pattern with 6 total experts , top-2 routing, 2 layers}
    \label{fig:6ex}
\end{figure}

\subsection*{S2 Ablation Study for Mixture of Expert Graph Transformer}\label{S2}
An ablation study is performed in order to understand the function of model's components and the impact that their configurations has on performance. The performance is assessed by using accuracy, precision,
recall and f1 score metrics, in addition, the specialization behavior
of the MoE layer is studied.
The dataset used in this section will be a portion of the complete one, containing 20000 signal events, 10000 background ttbar events and 10000 background singletop events.
The components studied are the following:
\begin{itemize}
    \item \textbf{Weight of the load balancing loss}: The factor by which the load balancing loss is multiplied, before it being added to the chosen training criterion
    \item \textbf{Learning rate}: Learning rate configuration, including the parameters of the scheduler function
    \item \textbf{Mixture of Experts layer configuration (number of experts, k-routing, expert size)}: Respectively: how many total experts the MoE layer has, how many experts will be active for each token, the size of the matrices that compose an expert
    \item \textbf{Attention layer configuration (hidden size, number of heads)}:Respectively: size of the graph's embedding and attention matrices, number of heads for the multi-head attention layer
    \item \textbf{Dropout probability}: Probability of dropout after the attention layer and inside the experts subnetwork
    \item \textbf{Number of layers}: How many encoder layers the model will be composed of.
\end{itemize}
Changing the weight of the load balancing loss does impact the experts utilization behaviour significantly. When setting a value of zero, the experts are not evenly utilized and their ability to specialize is compromised. This hyperparameter's ideal value is 1, which allows the model to even out experts utilization quickly and ensure that the experts can specialize, values higher than one do not grant any tangible improvements.

Adjusting the learning rate has a notable impact on performance results and experts specialization pattern, the higher the learning rate is, the more sharp and less noisy the specialization pattern becomes. On the other hand, having a learning rate too high introduces instability in the training process that results in performance loss.

The number of total experts, experts size and routing
choice have no measurable impact on the performance of the model,
however, they heavily contribute to the resulting specialization pattern. The experts tend to specialize on particle types, more specifically on groups of highly correlated particles, on this specific dataset, the model configuration that achieves the sharpest specialization on the experts is using top-2 routing over 6 total experts. Let's consider $k$, the number of experts simultaneously active, $n$ the number of total experts and $l$ groups of highly correlated particles in the graph data, the specialization pattern will tend to a configuration in which, for every particle type node, there is a $k-$plet of active experts, the same $k-$plet will be active on nodes belonging to the same group of correlated particles. The closest $n/k$ is to $l$, the sharper the specialization pattern becomes.\\
This behaviour is further tested with an additional ablation study done on the dataset. The graphs were truncated on the second to last node, leaving only jets and b-jets, 5 nodes in total for each graph in the dataset. In this case, the best performing model, with regard to specialization, has 4 total experts, two of them specialize on the three jets and the other two on b-jets \cref{fig:4ex}; whereas, the previously tested 6 experts configuration does not manage to specialize correctly. \cref{fig:6ex}\\

\noindent Changing the hidden size of the model does not result in significant differences, increasing the value of this hyperparameter too much leads to performance loss and run-to-run inconsistency. The number of heads also has no impact on performance or specialization behaviour.\\

\noindent Adding layers beyond the first did not result in noticeable performance increase.
Regarding the expert specialization behaviour, when testing the model with multiple layers, the MoE layers in deeper encoders saw a sharper and less noisy specialization pattern. This trend is seen in all of the MoE layer configurations tested.\\

\noindent Lowering dropout probability slightly improved the performance
of the model, reducing, however, the sharpness of the experts specialization patterns.
Also, the tendency of the specialization to be sharper at deeper layers reverses when setting the dropout probability to zero. When testing the model on the full dataset, the specialization capaibility did not suffer when lowering dropout probability, resulting only in a small accuracy boost without drawbacks.\\

\noindent Given the results of the ablation study it was decided to configure the model as follows:
\begin{itemize}
    \item \textbf{Hidden size}: 80
    \item \textbf{Number of heads}: 2
    \item \textbf{Number of experts and routing}: 6, top-2 routing
    \item \textbf{Dropout probability}: 0
    \item \textbf{Number of layers}: 2
\end{itemize}

\subsection*{S3 Training details} 

\begin{table}[H]
\centering
\begin{tabular}{|l|l|l|l|}
\hline
\textbf{Model} & \textbf{Architecture} & \textbf{Layers} & \textbf{Hidden Size} \\ \hline
\textbf{GCN} & \makecell{Graph Convolutional \\ Network (GCN)} & \makecell{2 Convolutional \\ Layers} & \makecell{Hidden channels: \\ 80 and 40} \\ \hline
\textbf{MLP} & \makecell{Multilayer \\ Perceptron} & \makecell{2 Hidden \\ Layers} & \makecell{Hidden layers: \\ 80 and 40} \\ \hline
\textbf{GT} & \makecell{Graph \\ Transformer} & \makecell{2 Transformer \\ Layers} & \makecell{Attention heads: 2; \\ Hidden size: 80} \\ \hline
\textbf{MGT} & \makecell{Mixture of Experts \\ Graph Transformer} & \makecell{2 Transformer Layers \\ with Mixture of \\ Experts Layer} & \makecell{Hidden size: 80; \\ Attention heads: 2; \\ 6 Experts \\ with size 20 each} \\ \hline
\end{tabular}
\caption{Training configurations for the models used in this study.}
\label{tab:trainingconfigurations}
\end{table}

The training configurations for the four models used in this study are detailed in \cref{tab:trainingconfigurations}. 

The parameters for the Mixture-of-Experts Graph Transformer (MGT) are selected based on the results of the ablation study presented in Section~S2, which determined the optimal configuration for this architecture. For the other models, their parameters are chosen to align as closely as possible with those of the MGT to ensure comparability while respecting their architectural differences.

The dataset for training consisted of 800,000 events, with equal representation of signal and background events (400,000 each). Background events included 200,000 examples of pair production of top quarks ($t\bar{t}$) and 200,000 examples of single top and $W$ boson production. 

The dataset was first split into 80\% training and 20\% test sets. The training subset was further divided into 80\% training and 20\% validation sets. 
All models were trained for up to 60 epochs and using a batch size of 500. Validation loss was monitored during training with early stopping. Cross-entropy loss was employed as the primary loss function, with an additional load balancing loss included for the MGT to optimize expert utilization.

\subsection*{S4 Analysis of the GCN results}

\begin{figure}[!ht]
    \centering
    \begin{subfigure}[b]{0.32\textwidth}
         \centering
         \includegraphics[width=\textwidth]{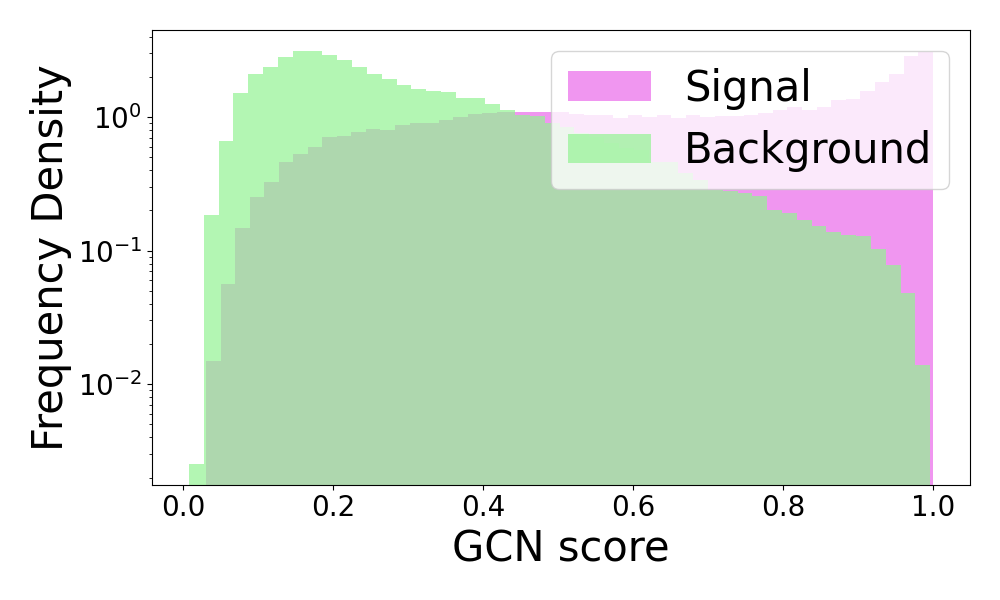}
         \caption{} \label{fig:outscore} 
         \end{subfigure}
         \begin{subfigure}[b]{0.32\textwidth}
         \centering
         \includegraphics[width=\textwidth]{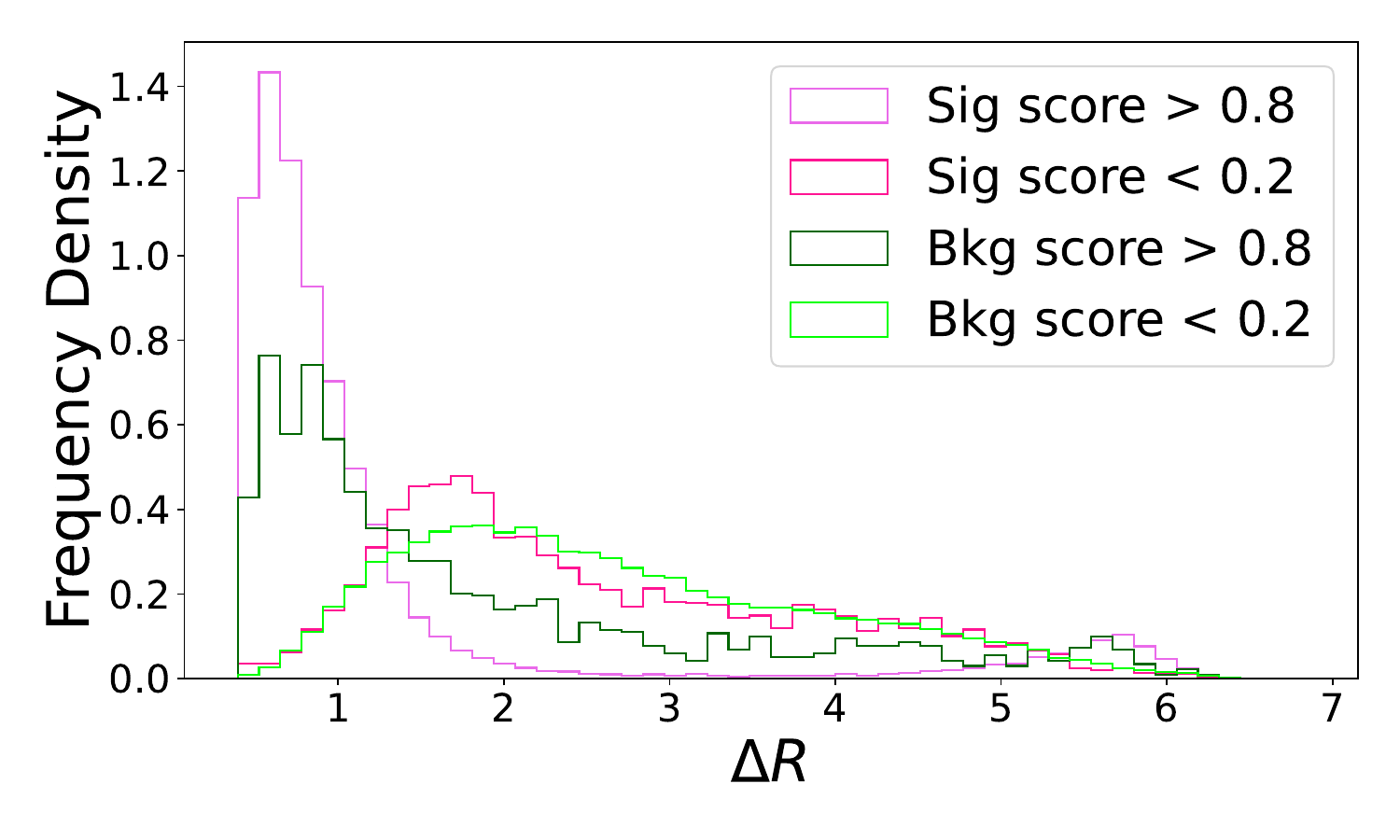}
         \caption{} \label{fig:DRbb} 
         \end{subfigure}
    \begin{subfigure}[b]{0.32\textwidth}
         \centering
         \includegraphics[width=\textwidth]{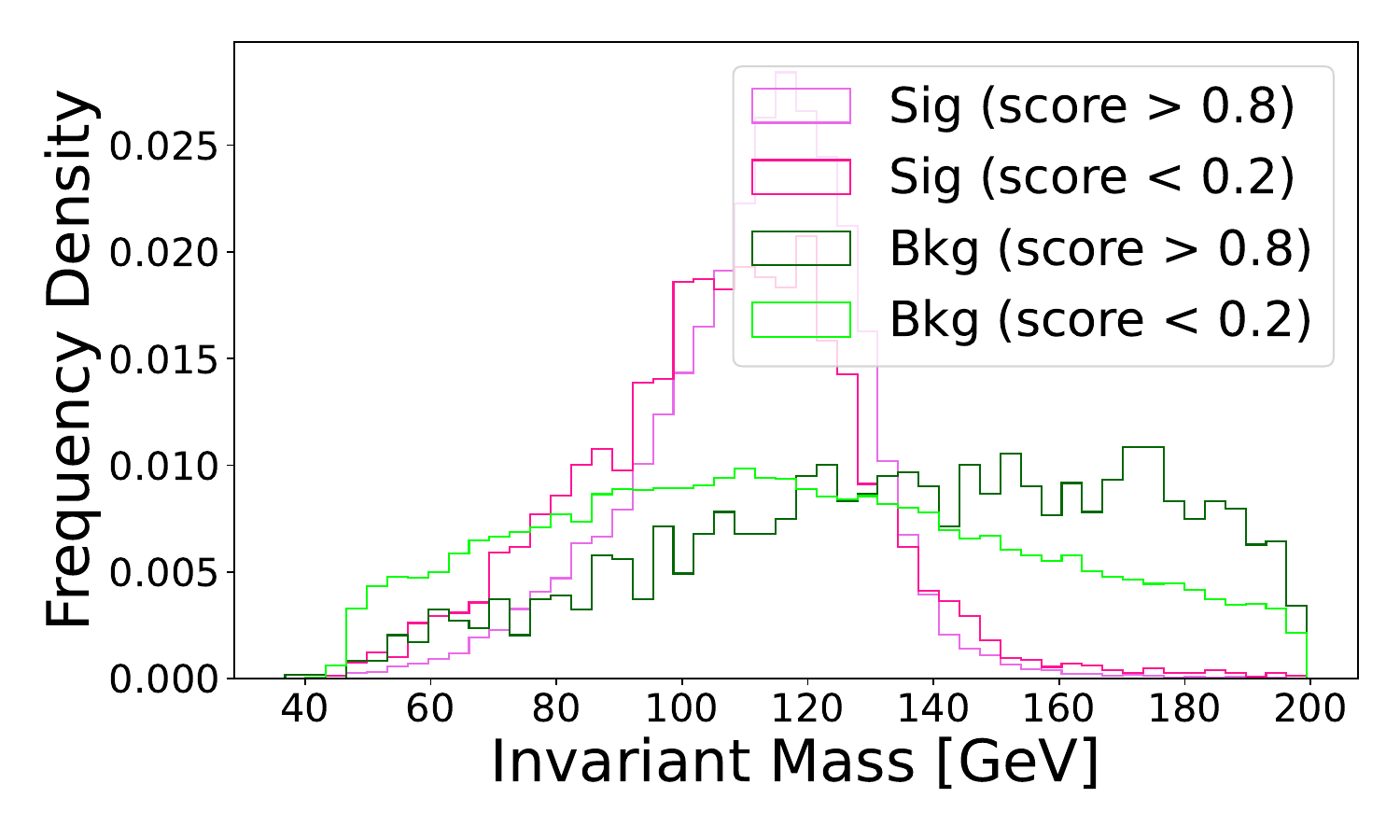}
         \caption{}
         \label{fig:mbb}
     \end{subfigure}
     \caption{(a) Output score distribution for the GCN. Distributions of complex variables built from correlation of features: (b) $\Delta R$ \particle{b1-b2} and, (c) invariant mass \particle{b1-b2}.}
\label{fig:GNNsusy}
\end{figure}

GCN results show good differentiation between signal and background, as shown in \cref{fig:GNNsusy}(a). The two SM sources are mixed and treated as equivalent, since it was found that there is no dependence of the results on the proportion of single and pair top-quark background. To interpret the learning process in a simplistic way, we tested the impact of systematically hiding certain input features from the datasets. As expected according to the kinematic features of the signal and background events, the main drop in performance was obtained when removing from the training either the $p_T$ features for \particle{b1}, \particle{b2}, or the features $p_T$ of the \particle{lepton} and the $E^{Miss}_{T}$. To also verify if the network effectively learns expected correlations within the graph, two key variables have been analysed depending on the output score: the spatial distance between \particle{b1} and \particle{b2} ($\Delta R$), and the invariant mass of the two particles. From a physics perspective, in signal events \particle{b1} and \particle{b2} are originated in the decay of the Higgs boson, a massive particle with mass of 125~GeV. Hence, it is expected that the two decay products \particle{b1} and \particle{b2} are close to each other spatially, and that their invariant mass is around that of the Higgs boson. In the case of SM background, such correlations are not expected. \cref{fig:GNNsusy} (b) and (c) show the distributions for $\Delta R$ and invariant mass for the \particle{b1-b2} system for signal and background in case of high ($>0.8$) and low ($<0.2$) output score: the GCN effectively separates better signal models with close-by \particle{b1},\particle{b2}, whilst background events with low score have larger spatial distance. In terms of the invariant mass, interpretability of the results is more difficult: however, we note how high-score background events (signal-like) tend to have higher invariant mass, which is less likely for signal event. 

Finally, to further verify how relevant are the correlations on the network learning process, an additional test was performed by splitting the signal datasets into three different groups, defined by the mass difference ($\Delta m$) of the parent and the dark matter particles, and performing a training run for each group separately. The three groups of samples were characterised by a $\Delta m =$~150~GeV (group 1), 200~GeV (group 2) and $>500 GeV$ (group 3), respectively. The higher the mass difference, the higher the boost of the final decay products, which in turn is expected to result in a better distinction of the SUSY signal with respect to the background events. This was confirmed by the tests: with GCN, the AUC was found to be 0.71~$\pm$~0.02 for group 1, 0.81~$\pm$~0.01 for group 2 and 0.98~$\pm$~0.01 for group 3. Using the MoE approach, AUC increases to 0.76~$\pm$~0.02 for group 1 and to 0.84~$\pm$~0.01 for group 2, remaining at 0.98~$\pm$~0.01 for group 3. This demonstrates that (1) the kinematic of the signal events is exploited as expected and (2) the MoE model provides better distinction between signal and background also in the most challenging physics case.

\bibliographystyle{unsrt}  
\bibliography{sample}  

\section*{Acknowledgements}
The work is partly funded by the European Union’s CHIST-ERA programme under grant agreement
CHIST-ERA-19-XAI-009 (MUCCA). DG is supported by
PNRR MUR project PE0000013-FAIR. SS is partly funded by Sapienza grants RM1221816BD028D6
(DESMOS) and RG123188B3EF6A80 (CENTS). The authors would like to thank Dr Joseph Carmignani, former research assistant at the University of Liverpool supported by the MUCCA project, for his invaluable work on the SUSY ATLAS analysis and publication of the input datasets.  

\section*{Author contribution statement}
Conceptualization and Design: A.D., C.S., D.G., M.D., S.G., S.S. Methodology: A.D., D.G., M.D., S.S., A.S. Dataset—collection and preparation: A.D., S.S. Experiments: A.D., D.G., A.S., L.W., S.V. Validation: A.D., C.S., L.W., M.D., S.S., S.V. Writing—original draft preparation: A.D., D.G., M.D., A.S. Writing—review and editing: A.D., C.S., D.G., M.D., S.G., S.S., L.W., S.V. Supervision: C.S., M.D., S.G., S.S

\end{document}